\documentclass[11pt, a4paper]{lumia}

\usepackage[sort&compress]{natbib}
\usepackage{xspace}

\theoremstyle{plain}

\newtheorem*{proposition*}{Proposition}

\theoremstyle{definition}

\theoremstyle{definition}

\newcommand{\E}{\mathbb{E}}

\def\eqref#1{equation~\ref{#1}}

\usepackage{graphicx}           
\usepackage{tikz}               
\usepackage[edges]{forest}      

\usepackage{url}                
\usepackage{xurl}               

\usepackage{array}              
\usepackage{longtable}          
\usepackage{multirow}           
\usepackage{makecell}           
\usepackage{ragged2e}           

\usepackage{mathtools}          
\usepackage{nicefrac}           

\usepackage{algorithm}          
\usepackage{algorithmicx}       
\usepackage{algpseudocode}      
\usepackage{listings}           

\usepackage{subcaption}         
\usepackage{wrapfig}            
\usepackage[export]{adjustbox}  

\usepackage[commandnameprefix=always]{changes}

\usepackage{xspace}             
\usepackage[normalem]{ulem}     
\usepackage{CJKutf8}            
\usepackage{wrapfig}

\usepackage[tikz]{bclogo}       
\usepackage[framemethod=tikz]{mdframed} 

\usepackage{lipsum}             
\usepackage{tocloft}            
\usepackage{afterpage}          
\usepackage{bbding}             
\usepackage{epigraph}           
\usepackage{minitoc}            
\usepackage{multicol}           
\usepackage{textgreek}          

\newcolumntype{P}[1]{>{\RaggedRight\arraybackslash}p{#1}}

\definecolor{uclablue}{RGB}{39, 116, 174}
\definecolor{bigaired}{RGB}{156, 0, 0}
\definecolor{myblue}{HTML}{598BE7}
\definecolor{mildblue}{RGB}{31,119,180}
\definecolor{sectionblue}{RGB}{70, 130, 180}
\definecolor{methodblue}{RGB}{0, 150, 136}
\definecolor{bgblue}{RGB}{245,243,253}
\definecolor{ttblue}{RGB}{91,194,224}
\definecolor{mygreen}{rgb}{0.64, 0.56, 0.88}
\definecolor{myyellow}{rgb}{0.68, 0.6, 0.1}
\definecolor{fancygreen}{rgb}{0.33, 0.68, 0.20}
\definecolor{salmon}{rgb}{0.94, 0.52, 0.49}
\definecolor{tablegreen}{rgb}{0.82, 0.94, 0.75}
\definecolor{tableblue}{rgb}{0.81, 0.90, 0.94}
\definecolor{tablered}{rgb}{0.97, 0.85, 0.85}
\definecolor{tableorange}{rgb}{0.96, 0.85, 0.81}
\definecolor{myorange}{rgb}{1.0, 0.49, 0.0}
\definecolor{tlgreen}{rgb}{0.33, 0.68, 0.20}
\definecolor{darkgreen}{RGB}{0,100,0}
\definecolor{darkred}{RGB}{200, 0, 0}
\definecolor{customyellow}{HTML}{FFFACD}
\definecolor{refinegreen}{RGB}{0, 128, 75}
\definecolor{scoregreen}{RGB}{34, 139, 34}
\definecolor{hidden-blue}{RGB}{194,232,247}
\definecolor{hidden-black}{RGB}{20,68,106}
\definecolor{yes}{HTML}{C6EFCE}
\definecolor{no}{HTML}{FFC7CE}
\definecolor{partial}{HTML}{FFEB9C}
\definecolor{external}{HTML}{D9E1F2}
\definecolor{hdr}{HTML}{F2F2F2}
\definecolor{GRPOrow}{gray}{0.96}
\definecolor{FlowRLrow}{RGB}{225,236,255}
\definecolor{FlowBlue}{RGB}{80,120,210}
\definecolor{GRPOGray}{gray}{0.35}

\hypersetup{
    colorlinks=true, 
    citecolor=uclablue, 
    linkcolor=bigaired,
    urlcolor=darkblue
}

\setlist[itemize]{leftmargin=20pt, noitemsep, topsep=0pt}

\NewDocumentCommand{\kaiyan}{mO{}}{\textcolor{purple}{\textsuperscript{\textit{kaiyan}}\textsf{\textbf{\small[#1]}}}}
\NewDocumentCommand{\yuxin}{mO{}}{\textcolor{cyan}{\textsuperscript{\textit{yuxin}}\textsf{\textbf{\small[#1]}}}}
\NewDocumentCommand{\bx}{mO{}}{\textcolor{green}{\textsuperscript{\textit{bx}}\textsf{\textbf{\small[#1]}}}}
\NewDocumentCommand{\at}{mO{}}{\textcolor{red}{\textsuperscript{\textit{AT}}\textsf{\textbf{\small[#1]}}}}
\NewDocumentCommand{\re}{mO{}}{\textcolor{blue}{\textsuperscript{\textit{RE}}\textsf{\textbf{\small[#1]}}}}
\NewDocumentCommand{\ybsun}{mO{}}{\textcolor{magenta}{\textsuperscript{\textit{youbang}}\textsf{\textbf{\small[#1]}}}}
\NewDocumentCommand{\runze}{mO{}}{\textcolor{orange}{\textsuperscript{\textit{runze}}\textsf{\textbf{\small[#1]}}}}
\NewDocumentCommand{\add}{mO{}}{\textcolor{darkgreen}{\textsuperscript{\textit{Maybe Consider Discuss}}\textsf{\textbf{[#1]}}}}

\newcommand{\cmark}{\textcolor{darkgreen}{\boldmath$\checkmark$}}
\newcommand{\xmark}{\textcolor{darkred}{\boldmath$\times$}}

\newcommand{\MethodName}{ConceptLM}

\newenvironment{itemize*}%
 {\leftmargini=10pt\begin{itemize}%
  \setlength{\itemsep}{0pt}%
  \setlength{\parskip}{0pt}%
  }%
 {\end{itemize}}

\newenvironment{enumerate*}%
 {\begin{enumerate}%
  \setlength{\itemsep}{0pt}%
  \setlength{\parskip}{0pt}}%
 {\end{enumerate}}

\newcommand{\cellstatus}[1]{%
  \begingroup
  \StrTrim{#1}[\statusval]%
  \IfStrEq{\statusval}{Yes}{\cellcolor{yes}\cmark}{}%
  \IfStrEq{\statusval}{No}{\cellcolor{no}\xmark}{}%
  \IfBeginWith{\statusval}{Yes (}{\cellcolor{yes}\cmark~\textit{\statusval\unskip}}{}%
  \IfStrEq{\statusval}{Partial}{\cellcolor{partial}\textbf{Partial}}{}%
  \IfStrEq{\statusval}{External}{\cellcolor{external}\textbf{External}}{}%
  \endgroup
}

\newtcolorbox{myboxi}[1][]{
  breakable,
  title=#1,
  colback=red!5,
  colbacktitle=red!5,
  coltitle=black,
  fonttitle=\bfseries,
  bottomrule=0pt,
  toprule=0pt,
  leftrule=2pt,
  rightrule=2pt,
  titlerule=0pt,
  arc=0pt,
  outer arc=0pt,
  colframe=red,
}

\newtcolorbox{myboxnote}[1][]{
  breakable,
  title=#1,
  colback=orange!0,
  colbacktitle=orange!0,
  coltitle=black,
  fonttitle=\bfseries,
  bottomrule=0pt,
  toprule=0pt,
  leftrule=2pt,
  rightrule=2pt,
  titlerule=0pt,
  arc=0pt,
  outer arc=0pt,
  colframe=orange,
}

\newtcolorbox{myboxii}[1][]{
  breakable,
  freelance,
  title=#1,
  colback=white,
  colbacktitle=white,
  coltitle=black,
  fonttitle=\bfseries,
  bottomrule=0pt,
  boxrule=0pt,
  colframe=white,
  overlay unbroken and first={
  \draw[red!75!black,line width=3pt]
    ([xshift=5pt]frame.north west) -- 
    (frame.north west) -- 
    (frame.south west);
  \draw[red!75!black,line width=3pt]
    ([xshift=-5pt]frame.north east) -- 
    (frame.north east) -- 
    (frame.south east);
  },
  overlay unbroken app={
  \draw[red!75!black,line width=3pt,line cap=rect]
    (frame.south west) -- 
    ([xshift=5pt]frame.south west);
  \draw[red!75!black,line width=3pt,line cap=rect]
    (frame.south east) -- 
    ([xshift=-5pt]frame.south east);
  },
  overlay middle and last={
  \draw[red!75!black,line width=3pt]
    (frame.north west) -- 
    (frame.south west);
  \draw[red!75!black,line width=3pt]
    (frame.north east) -- 
    (frame.south east);
  },
  overlay last app={
  \draw[red!75!black,line width=3pt,line cap=rect]
    (frame.south west) --
    ([xshift=5pt]frame.south west);
  \draw[red!75!black,line width=3pt,line cap=rect]
    (frame.south east) --
    ([xshift=-5pt]frame.south east);
  },
}

\tcbset{
  takeawaysbox/.style={
    title=Takeaways,
    colback=lightblue!80,
    colframe=black,
    fonttitle=\bfseries\small,
    coltitle=white,
    colbacktitle=black,
    enhanced,
    attach boxed title to top left={xshift=2.5mm,yshift=-2.5mm},
    boxed title style={rounded corners, size=small, colframe=black, colback=black},
    width=\linewidth,
    arc=3.5mm
  }
}

\mdfdefinestyle{mystyle}{%
  rightline=true,
  innerleftmargin=10,
  innerrightmargin=10,
  outerlinewidth=3pt,
  topline=false,
  rightline=true,
  bottomline=false,
  skipabove=\topsep,
  skipbelow=\topsep
}

\tikzset{%
    every node/.style={font=\tiny},
    parent/.style =          {align=center,text width=2cm,rounded corners=3pt, line width=0.3mm, fill=gray!10,draw=gray!80},
    child/.style =           {align=center,text width=2.0cm,rounded corners=3pt, fill=blue!10,draw=blue!80,line width=0.3mm},
    grandchild/.style =      {align=center,text width=2cm,rounded corners=3pt},
    greatgrandchild/.style = {align=center,text width=1.5cm,rounded corners=3pt},
    greatgrandchild2/.style = {align=center,text width=1.5cm,rounded corners=3pt},    
    referenceblock/.style =  {align=center,text width=1.5cm,rounded corners=2pt},
    pretrain/.style =           {align=center,text width=2.0cm,rounded corners=3pt, fill=blue!10,draw=blue!80,line width=0.3mm},   
    pretrain_work/.style =           {align=center, text width=8.5cm,rounded corners=3pt, fill=blue!10,draw=blue!0,line width=0.3mm},  
    template/.style =           {align=center,text width=2.0cm,rounded corners=3pt, fill=red!10,draw=red!80,line width=0.3mm},   
    template_work/.style =           {align=center,text width=8.5cm,rounded corners=3pt, fill=red!10,draw=red!0,line width=0.3mm},    
    answer/.style =           {align=center,text width=2.0cm,rounded corners=3pt, fill= cyan!10,draw= cyan!80,line width=0.3mm},   
    answer_work/.style =           {align=center,text width=8.5cm,rounded corners=3pt, fill= cyan!10,draw= cyan!0,line width=0.3mm},      
    multiple/.style =           {align=center,text width=2.0cm,rounded corners=3pt, fill= orange!10,draw= orange!80,line width=0.3mm},   
    multiple_work/.style =           {align=center,text width=8.5cm,rounded corners=3pt, fill= orange!10,draw= orange!0,line width=0.3mm},        
    tuning/.style =           {align=center,text width=2.0cm,rounded corners=3pt, fill= magenta!10,draw= magenta!80,line width=0.3mm},   
    tuning_work/.style =           {align=center,text width=8.5cm,rounded corners=3pt, fill= magenta!10,draw= magenta!0,line width=0.3mm},          
}

\newcommand{\lstbg}[3][0pt]{{\fboxsep#1\colorbox{#2}{\strut #3}}}

\lstdefinelanguage{diff}{
  basicstyle=\ttfamily\small,
  morecomment=[f][\lstbg{red!20}]-,
  morecomment=[f][\lstbg{green!20}]+,
}

\lstdefinelanguage{diffpython}{
  language=diff,
  morekeywords={def, if, else, for, while, return, import, from, as, class, with, try, except, finally, raise, lambda, and, or, not, in, is, None, True, False},
  morecomment=[l]{\#},
  morestring=[b]",
  morestring=[b]',
}

\setheadertext{LUMIA Lab}

\correspondingemail{\emailicon \{liuyl03181, lin.zhouhan\} @gmail.com \quad $^\ddagger$ Corresponding Author. }

\githublink{Code and Models: https://github.com/LUMIA-Group/ConceptLM}

\renewcommand{\today}{2026-02-10}

\title{Next Concept Prediction in Discrete Latent Space Leads to Stronger Language Models}
\setheadertitle{NCP Leads to Stronger Language Models}

\author{%
  Yuliang Liu$^{1,2,5}$, Yunchong Song$^{2}$, Yixuan Wang$^{1,2,5}$, Kewen Ge$^{1}$, Alex Lamb$^{4}$, Qipeng Guo$^{2}$, Kai Chen$^{2}$, Bowen Zhou$^{2,3}$, Zhouhan Lin$^{1,2\ddagger}$ \\
  $^1$ LUMIA Lab, School of Artificial Intelligence, Shanghai Jiao Tong University \\
  $^2$ Shanghai AI Laboratory \\
  $^3$ Department of Electronic Engineering, Tsinghua University \\
  $^4$ College of Artificial Intelligence, Tsinghua University \\
  $^5$ Shanghai Innovation Institute

}

\begin{document}

\begin{abstract}

We propose Next Concept Prediction (NCP), a generative pretraining paradigm built on top of Next Token Prediction (NTP). NCP predicts discrete concepts that span multiple tokens, thereby forming a more challenging pretraining objective. Our model, ConceptLM, quantizes hidden states using Vector Quantization and constructs a concept vocabulary. It leverages both NCP and NTP to drive parameter updates and generates a concept to guide the generation of the following tokens. We train ConceptLM from scratch at scales ranging from 70M to 1.5B parameters with up to 300B training data, including Pythia and GPT-2 backbones. Results on 13 benchmarks show that NCP yields consistent performance gains over traditional token-level models. Furthermore, continual pretraining experiments on an 8B-parameter Llama model indicate that NCP can further improve an NTP-trained model. Our analysis suggests that NCP leads to more powerful language models by introducing a harder pretraining task, providing a promising path toward better language modeling.

\end{abstract}

\maketitle


\section{Introduction}

\begin{figure}[!b]
    \centering
    \begin{subfigure}{0.31\textwidth}
        \centering
        \includegraphics[width=\linewidth]{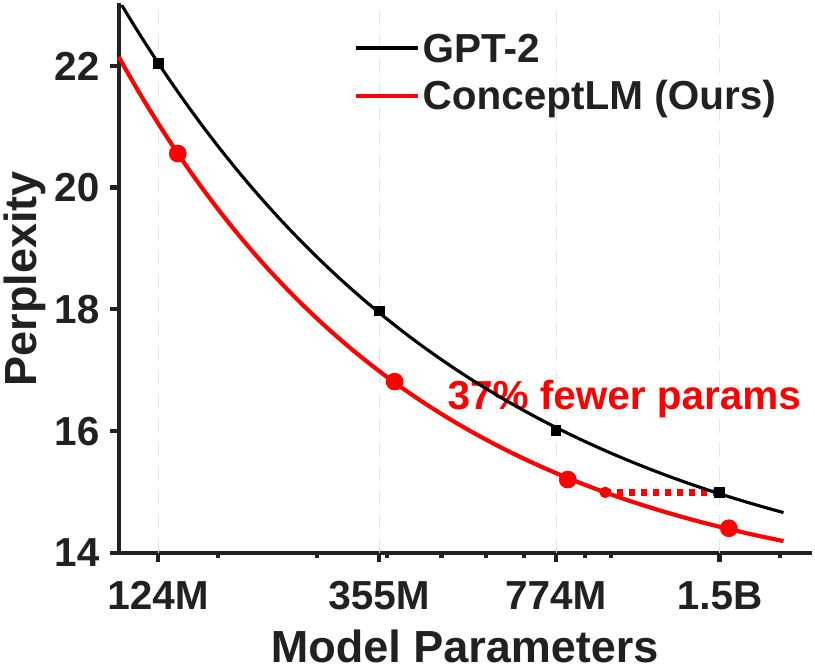}
    \end{subfigure}
    \hspace{0.01\textwidth}
    \begin{subfigure}{0.31\textwidth}
        \centering
        \includegraphics[width=\linewidth]{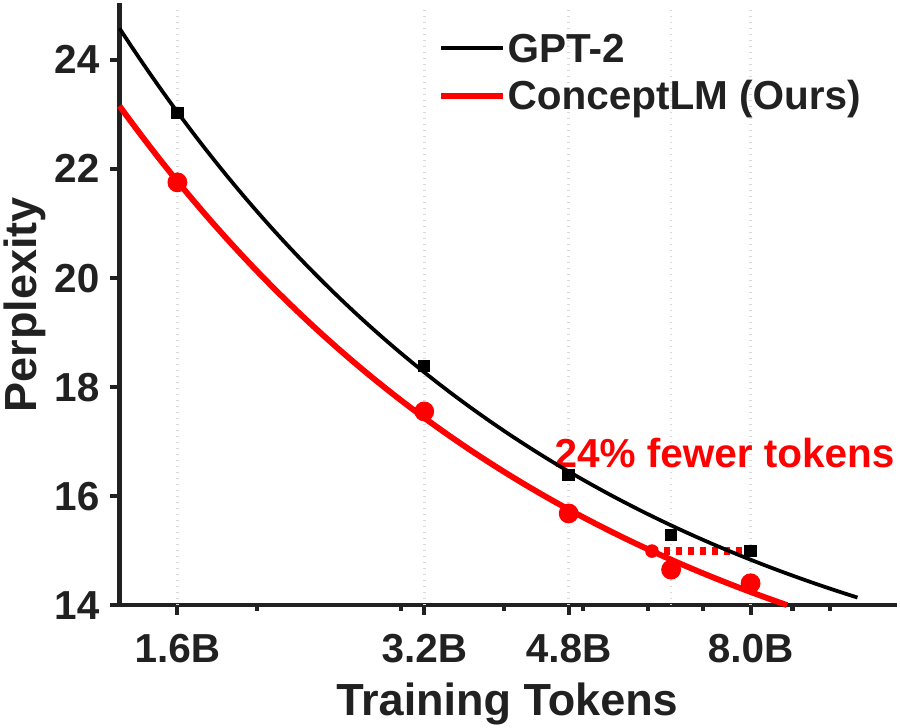}
    \end{subfigure}
    \hspace{0.01\textwidth}
    \begin{subfigure}{0.31\textwidth}
        \centering
        \includegraphics[width=\linewidth]{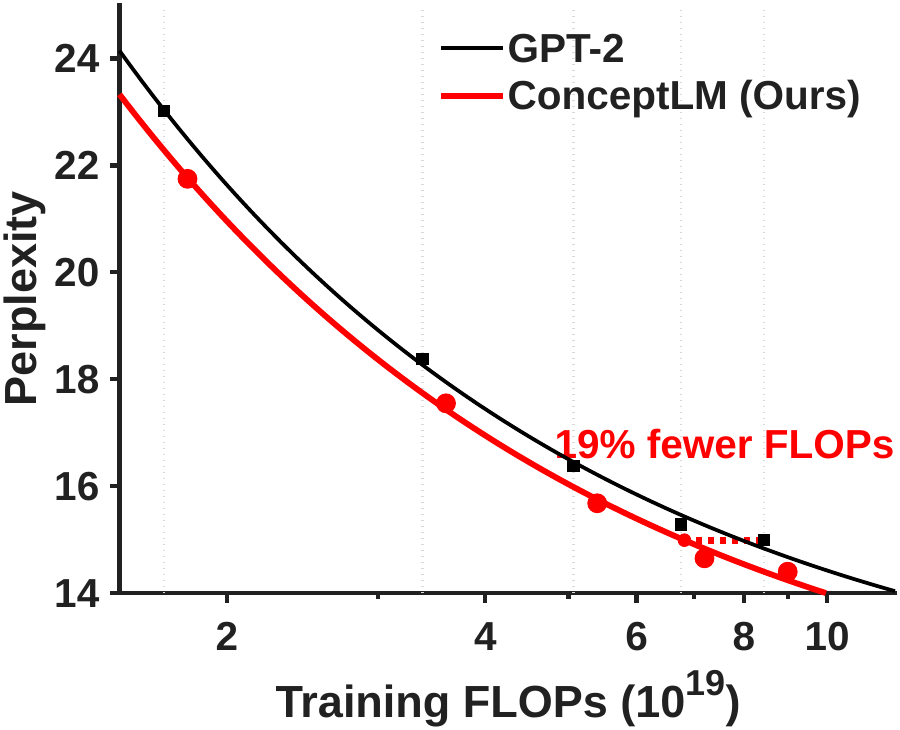}
    \end{subfigure}

    \caption{Scaling performance comparison of ConceptLM against traditional token-level models across three key dimensions: model parameters, training tokens, and total FLOPs. The results demonstrate that ConceptLM exhibits superior scalability.}
    \label{fig:scaling_triple}
\end{figure}

The remarkable success of modern AI is largely driven by the generative pretraining paradigm, where intelligence emerges from training on large amounts of data to predict future units~\citep{brown2020languagemodelsfewshotlearners, ho2020denoisingdiffusionprobabilisticmodels}. However, as models scale up, low-level learning objectives cannot fully utilize model capacity, bottlenecking AI models' potential~\citep{lecun2022path}. 
The evolution in the abstract level of prediction is pratically evident across domains: modern computer vision methods increasingly emphasize latent that transcend pixel-level details in favor of high-level abstractions~\citep{tian2024visualautoregressivemodelingscalable}; sequential decision-making in modern RL and robotics treats decision tasks as trajectory-level sequence modeling rather than independent, step-wise action mapping~\citep{chen2021decisiontransformerreinforcementlearning, chi2024diffusionpolicyvisuomotorpolicy}; and multimodal methods, like LCM ~\citep{lcmteam2024largeconceptmodelslanguage}, elevate the predictive target to a fixed embedding space, focusing on predicting whole sentences.
Joint Embedding Predictive Architectures (JEPA)~\citep{assran2023selfsupervisedlearningimagesjointembedding} demonstrates the efficacy of learning through a latent space rather than from raw inputs, and reveals the rationale for learning in a high-level space from a systematic perspective.
Yet LLMs, despite reaching trillion-parameter scale, remain constrained to token-level prediction~\citep{grattafiori2024llama3herdmodels, yang2025qwen3technicalreport}, which may no longer be sufficient to unlock their full power.

Beyond token-level modeling, several studies have explored language modeling at other semantic levels. For instance, the Hourglass Transformer~\citep{hourglass} and ContextLM~\citep{dai2025contextlevellanguagemodelinglearning} compress sequences to learn high-level dense representations, while H-Net~\citep{hwang2025dynamicchunkingendtoendhierarchical} and BLT~\citep{pagnoni2024bytelatenttransformerpatches} target lower-level modeling, introduce dynamic segmentation, and are thus tokenizer-free. Despite modeling hierarchical representation, these models remain reliant on standard next token prediction or next byte prediction training objectives. Multi-Token Prediction~\citep{gloeckle2024betterfasterlarge} takes a step toward a harder training task by predicting multiple tokens simultaneously, yet it remains confined to the token space. Directly predicting a set of tokens, rather than a single one, increases the learning difficulty combinatorially.

We propose Next Concept Prediction (NCP), a concept-level pretraining objective operating within a discrete latent space, as the core of the ConceptLM framework. ConceptLM integrates concept-level and token-level prediction by utilizing Vector Quantization (VQ) to project continuous concept representations, which are aggregated from token-level hidden states, into finite learnable codebooks. This establishes a discrete concept vocabulary, thereby providing explicit targets for concept-level prediction. The model then predicts next tokens conditioned on the corresponding predicted concept, rather than ground-truth future concepts, to prevent information leakage. By jointly optimizing NCP and Next-Token Prediction (NTP), our model implicitly implements planning before generation. Our main contributions include:
\begin{itemize}
\item \textbf{Models with NCP scale up better:} We train GPT-2 and Pythia backbones from scratch, scaling up to 1.5B parameters for GPT-2 with 8B tokens, and 410M for Pythia with 300B tokens, demonstrating that ConceptLM achieves significant improvements in various language modeling and downstream tasks compared with token-level baselines. We provide the scaling performance from three perspectives in Figure~\ref{fig:scaling_triple}.
\item \textbf{NCP can enhance existing LLMs:} By using only 9.6B tokens for continual pretraining, we successfully extend the existing Llama model into a concept-level model, achieving an average performance gain of 0.4 on the 8B-parameter scale, compared with its token-level counterpart across 4 benchmarks.
\item \textbf{NCP improves long-range modeling:} Detailed ablation studies confirm that NCP drives the performance gains. Furthermore, our analysis reveals that NCP encourages the model to focus on long-range dependencies, enhancing its global modeling capability, and that NCP makes the models scaling behavior better.
\end{itemize}

\section{ConceptLM Architecture}

\begin{figure}[htb]
    \centering
    \includegraphics[width=\linewidth]{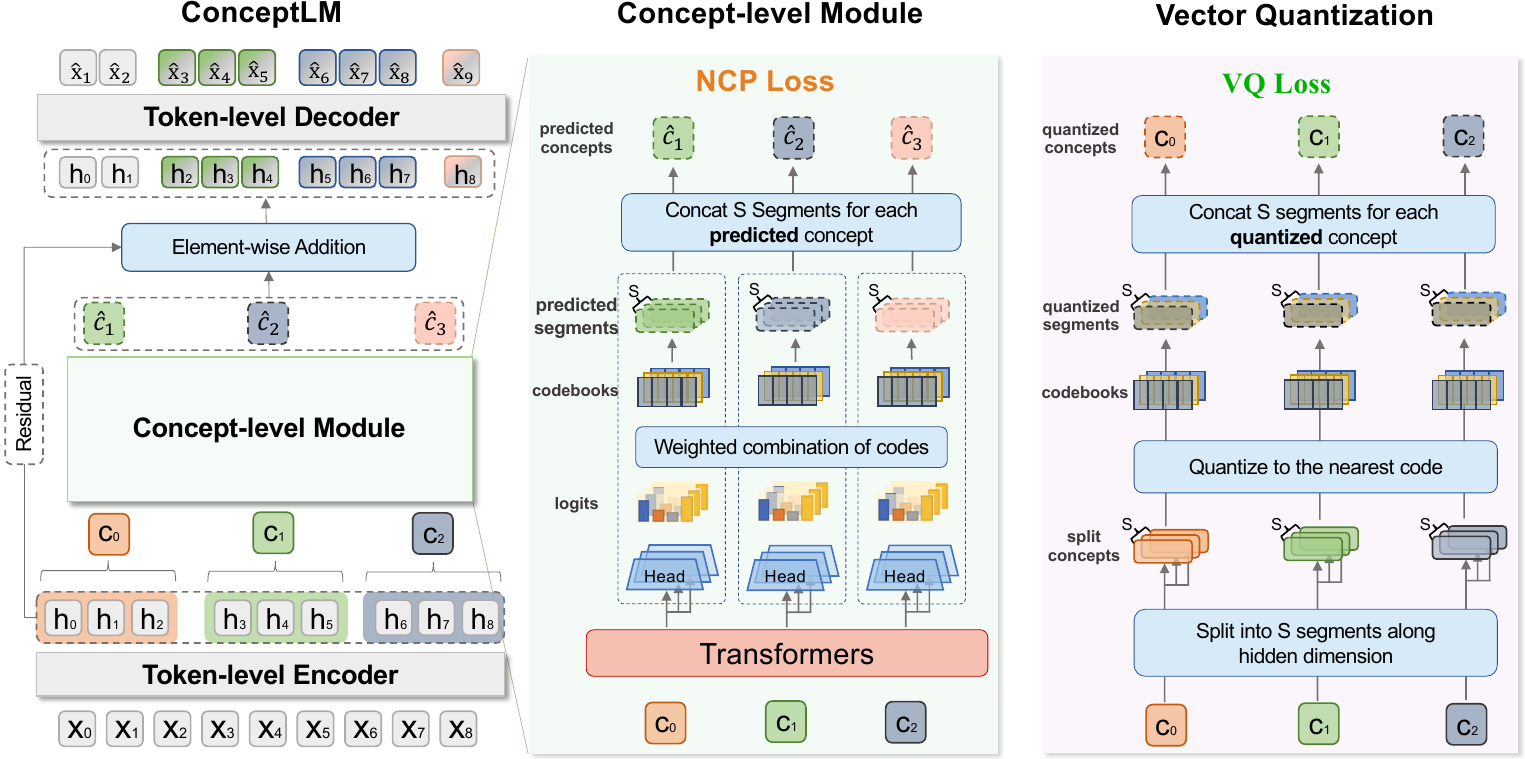}
\caption{Overview of \MethodName. The figure is organized into three components: the left one shows the ConceptLM workflow; the central part shows the concept prediction process; and the right one shows the segmented VQ strategy used to train the concept vocabulary. We only quantize the concept-level representations to train the codebooks in the training phase.}
\label{fig:main_arc}
\end{figure}

In this section, we introduce \MethodName\ architecture, including the Token-level Encoder, the Concept-level Module, and the Token-level Decoder. 
\MethodName\ construct a concept-level discrete latent space $\mathcal{D}_c$ that resides on the continuous latent space $\mathcal{H}_c$, and then perform concept-level prediction on the concept space, thereby driving the token prediction of the model with conceptual supervision.
Figure~\ref{fig:main_arc} shows the overview of our model.

\subsection{Token-level Encoder}

Given a raw token sequence, the Token-level Encoder first encodes tokens into continuous token-level 
\begin{wrapfigure}{r}{0.48\linewidth}
    \centering
    \includegraphics[width=0.8\linewidth]{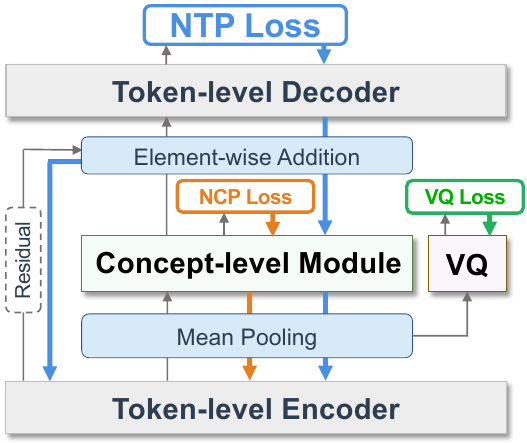}
\caption{Forward and backward of \MethodName.}
\label{fig:forward_backward}
  \vspace{-0.5em}  
\end{wrapfigure}
hidden states $h \in \mathcal{H}$, which serve as the foundational intermediate representations. These hidden states are designed for three purposes: (1) providing token-level hidden states for the Token-level Decoder to facilitate predicting the next token; (2) serving as the source for compression into concept-level hidden states for predicting the next concept; and (3) being projected into the discrete latent space for concept-level vocabulary optimization.

\subsection{Concept-level Module}

With the continuous token-level hidden states, we derive concept-level hidden states by aggregating adjacent token-level ones with a mean-pooling function $f_c: \mathcal{H} \to \mathcal{H}_c$. The length of the concept-level sequence is reduced to $T/k$, where $k$ denotes the compression factor. Then, the model predicts the next concept over a concept vocabulary within a discrete latent space defined by a Quantizer operating on the concept hidden states.

\paragraph{Construct the Concept-level Vocabulary: } We employ Vector Quantization (VQ)~\citep{Vector_Quantization} to transform the continuous concept-level states into a discrete and predictable space. The VQ maps continuous latent states into a finite set of representative embeddings, denoted as $f: \mathcal{H}_c \to \mathcal{D}_c$.

To enhance the representational capacity of VQ, we adopt a product quantization strategy~\citep{jegou2010product}. Specifically, each continuous concept representation is partitioned into $S$ independent segments along the feature dimension. Each segment is then quantized using a separate codebook containing $N$ learnable entries $\{e^s_1, e^s_2, \dots, e^s_N\}$ for the $s$-th segment. This results in a total discrete representation space of size $N^S$, significantly increasing the expressivity of this discrete space while maintaining a small codebook size, with a near-zero cost in terms of quantization latency.

\paragraph{Predict the Next Concept: }

To predict the subsequent concept, the Concept-level Module employs Transformer layers to process the historical continuous concept hidden states $\mathbf{c}_{<t/k} \in \mathcal{H}_c$. Subsequently, $S$ independent prediction heads are deployed to estimate the multi-segment discrete representation of the target concept. Specifically, each head performs a linear projection from the hidden state dimension to the codebook size $N$, producing a logits distribution over the entries of its corresponding segment. Formally, for the $s$-th segment, we predicts
\[
w^s_n = P_{\theta^s_c}(h_{<t/c}), \quad n = 1, \dots, N,
\]
where $N$ is the number of codebook entries for this segment, and $\theta^s_c$ denotes the parameters of the shared Transformer layers with segment-specific prediction head.

Rather than performing hard sampling, which would introduce non-differentiability, the predicted segment representation is derived as a weighted sum of the codebook entries based on these predicted logits, noted by:
\[
\hat{c}^s = \sum_{n=1}^{N} w^s_n \, e^s_n.
\]
The final predicted conceptual representation $\hat{c}$ is the concatenation of all segment-wise predictions:

$$\hat{c} = \text{concat}(\hat{c}^1, \hat{c}^2, \dots, \hat{c}^S).$$

\subsection{Token-level Decoder}

Finally, the Decoder performs autoregressive generation to predict the next token. To align with the number of corresponding token hidden states of each concept, the predicted concept representation $\mathbf{\hat{c}}$ is broadcast $k$ times to match the token-level hidden states. This broadcast concept representation is then integrated with the token-level hidden states via element-wise addition. 

\textbf{Prevent Information Leakage:} Since each concept spans multiple tokens, predicting the next concept requires access to the complete Encoder hidden states $h$ of the previous concept. This leads to a misalignment in which the first $k-1$ tokens of a sequence lack a predicted concept. To prevent information leakage, we shift the broadcast concept sequence forward by $k-1$ positions before the element-wise addition, adding zero vectors to the first $k-1$ tokens.

This element-wise addition can be simply described by:

$$\tilde{h}_t = \begin{cases} 
h_t + \mathbf{0}, & 0 \le t < k-1 \\
h_t + \mathbf{\hat{c}}_{\lfloor t/(k-1) \rfloor}, & k-1 \le t < T 
\end{cases}$$

The Token-level Decoder with parameter $\theta$ predicts the next token based on the fused representation, formally:

\begin{equation*}
p(x_{t+1} \mid x_{\le t}) = P_{\theta}(\tilde{h}_{<t})\ .
\end{equation*}

\section{ConceptLM Training}
\label{subsec:model_training}

We delineate the training objectives of \MethodName, including $\mathcal{L}_{\text{VQ}}$ that construct discrete concept vocabulary; $\mathcal{L}_{\text{NCP}}$ that supervise the next concept prediction; and $\mathcal{L}_{\text{NTP}}$ that supervise the next token prediction. We show the forward and backward details in Figure~\ref{fig:forward_backward}.

\paragraph{The Supervision Signal to Construct Concept Vocabulary:}
$\mathcal{L}_{\mathrm{VQ}}$ is strictly confined to the Quantizer module without affecting the model's representation learning. By isolating this loss, we ensure the codebook entries effectively track the latent space without imposing quantization-specific constraints back onto the hidden representations.
Formally, the VQ loss is defined as:
\[
\mathcal{L}_{\text{VQ}} = \| \text{sg}(c) - d \|^2_2 + \beta \| c - \text{sg}(d) \|^2_2,
\]
where $c \in \mathcal{H}_c$ denotes the input concept-level hidden state, $d \in \mathcal{D}_c$ is the corresponding quantized embedding, and $\text{sg}(\cdot)$ denotes the stop-gradient operator.

\paragraph{The Supervision Signal to Predict Next Concept:} Building upon this discrete space, $\mathcal{L}_{\text{NCP}}$ supervises the NCP LM to predict next concept. To supervise this predictive process, we employ a Mean Squared Error (MSE) loss that aligns the predicted concept $\hat{c}$ with the ground-truth continuous latent states $h_c$. It is calculated over all predicted time steps:

$$\mathcal{L}_{\text{NCP}} = \frac{1}{M-1} \sum_{t=1}^{M-1} \| \hat{c}_t - c_t \|^2_2, M=T/k.$$

By regressing to a weighted combination of codebook representations, the $\mathcal{L}_{\text{NCP}}$ effectively supervises the next concept prediction while avoiding unconstrained optimization. 
The context prediction gradients propagate back to the Encoder, forcing the Encoder to capture high-level semantic and long-range dependencies that are essential for predicting future concepts, rather than merely focusing on tokens.

\paragraph{The Supervision Signal to Predict Next Token: } 
With the fused token-level representation $\tilde{h}$, the resulting token-level supervision can be formulated by:

\begin{equation*}
\begin{aligned}
\mathcal{L}_{\text{NTP}} &= -\sum_{t=0}^{T-1} \log P(x_{t+1} \mid x_{\leq t}, \hat{c}_{\leq \lfloor t/(k-1) \rfloor}) \\
&= -\sum_{t=1}^{T-1} \log P(x_{t+1} \mid \tilde{h}_t)
\end{aligned}
\end{equation*}

where the conditional dependency $(x_{\leq t}, \hat{c}_{\le \lfloor t/(k-1) \rfloor})$ is instantiated by the integrated representation $\tilde{h}_t$. 
$\mathcal{L}_{\mathrm{CE}}$ provides dense, end-to-end supervision. It affects all trainable parameters through backpropagation, ensuring the model maintains its fundamental language modeling capabilities.

\paragraph{Total Training Objective of \MethodName: } Finally, we train the model end-to-end with the total training objective:

\[
\mathcal{L} = \mathcal{L_\mathrm{NTP}} + \mathcal{L_\mathrm{NCP}} + \mathcal{L_\mathrm{VQ}}.
\]

\section{Experiments}

To evaluate \MethodName, we conduct several experiments across various model backbones, parameter scales, and training stages: training from scratch and continual pre-training.
Section~\ref{sub_sec:exp_set} details the general experimental setup and the configurations for \MethodName. Section~\ref{sub_sec:scaling_exp} shows the scaling experiments for the GPT-2~\citep{radford2019language} backbone, from 160M to 1.5B parameters with 8B tokens, and for the Pythia~\citep{Pythia} backbone, from 70M to 410M parameters with 300B tokens. 
Furthermore, we continually pre-train the existing token-level Llama-3.1-8B to extend it into concept models in Section~\ref{sub_sec:8b_models}.

\subsection{Experimental Setup}
\label{sub_sec:exp_set}

\textbf{Models and Datasets}: For scaling experiments, we use GPT-2 and Pythia series models as backbones, and train them using OpenWebText~\citep{Gokaslan2019OpenWeb} and the Pile~\citep{gao2020pile800gbdatasetdiverse} dataset from scratch, respectively. For 8B models, we use Llama-3.1-8B as the base model and perform continual pre-training using the LongCtxEng dataset~\citep{10.5555/3692070.3692634}. We follow the official hyperparameters in the scaling experiments; we list the detailed experiment settings in Appendix~\ref{App:exp_setting}.

\textbf{Configurations for \MethodName}: For the VQ, to avoid representation collapse, we draw inspiration from SimVQ~\citep{Zhu_2025_ICCV} by introducing a two-layer MLP with ReLU activation to transform the codebook embeddings. We set the number of segments $S=\mathrm{the \ head \ numbers}$ in each model and use a small $N$ of 64. And we use a two-layer Transformer backbone for concept-level processing with a chunk size of 4 ($k$=4) in all main experiments. 
We insert the Concept-level Module before the first layer for GPT-2 models and after the first layer for Pythia and Llama models. A more in-depth discussion regarding the insert depth and VQ is provided in Appendix~\ref{app:conceptlm_setup} and Appendix~\ref{app:VQ_discussion}.

\subsection{Pre-train ConceptLM from Scratch}
\label{sub_sec:scaling_exp}

We evaluate our method from a scaling perspective using GPT-2 and Pythia backbones. For each trained model, we employ a diverse suite of benchmarks: 9 downstream tasks (Lambada OpenAI/Standard~\citep{paperno2016lambadadatasetwordprediction}, ARC-Easy/Challenge~\citep{clark2018thinksolvedquestionanswering}, WinoGrande~\citep{sakaguchi2019winograndeadversarialwinogradschema}, PIQA~\citep{bisk2019piqareasoningphysicalcommonsense}, HellaSwag~\citep{zellers2019hellaswagmachinereallyfinish}, SciQ~\citep{welbl2017crowdsourcingmultiplechoicescience}, RACE~\citep{lai2017racelargescalereadingcomprehension}) and 4 language modeling tasks, covering Lambada OpenAI/Standard, Wikitext~\citep{merity2016pointersentinelmixturemodels}, and the original training distributions~\footnote{All benchmarks except the Pile and OpenWebText are evaluated using the \href{https://github.com/EleutherAI/lm-evaluation-harness}{Language Model Evaluation Harness} V-0.4.4.}. Detailed benchmark descriptions are available in Appendix~\ref{App:downstream_tasks}.

\begin{table}[ht]
\centering
\small 
\setlength{\tabcolsep}{3pt} 
\caption{Language Modeling results for GPT-2 and Pythia models. O/P denotes the OpenWebText results for GPT-2 models and Pile results for Pythia models. Lambada-O and Lambada-S denote the Lambada OpenAI and Lambada Standard results. We only annotated the model size for the base model in each group of experiments.}
\begin{tabular}{lccccc}
\toprule
\textbf{Model} & \textbf{O/P} & \textbf{Wiki} & \textbf{Lambada-O} & \textbf{Lambada-S} & \textbf{Avg PPL} $\downarrow$ \\
\midrule

GPT2-124M & 22.04 & 43.99 & 69.29 & 272.43 & 101.94 \\
GPT2-PM & 21.22 & 42.15 & 60.05 & 273.12 & 99.14 \\
ContextLM & 20.68 & 41.45 & 55.22 & 231.41 & 87.19 \\
\rowcolor[HTML]{F2F3F5} \textbf{\MethodName} & \textbf{20.56} & \textbf{40.86} & \textbf{51.83} & \textbf{225.74} & \textbf{84.75\textcolor{green!60!black}{\scriptsize{\({\downarrow17.19}\)}}} \\

\addlinespace[0.3em]

GPT2-355M & 17.97 & 34.24 & 35.54 & 122.54 & 52.57 \\
GPT2-PM & 17.67 & 33.59 & 32.91 & 126.03 & 52.55 \\
ContextLM & 17.03 & 32.08 & 28.06 & 95.06 & 43.06 \\
\rowcolor[HTML]{F2F3F5} \textbf{\MethodName} & \textbf{16.81} & \textbf{31.53} & \textbf{26.96} & \textbf{86.75} & \textbf{40.51\textcolor{green!60!black}{\scriptsize{\({\downarrow12.06}\)}}} \\

\addlinespace[0.3em]

GPT2-774M & 16.01 & 30.03 & 24.58 & 74.76 & 36.35 \\
GPT2-PM & 15.89 & 29.71 & 23.50 & 62.45 & 32.89 \\
ContextLM & 15.41 & 28.82 & 20.89 & 62.09 & 31.80 \\
\rowcolor[HTML]{F2F3F5} \textbf{\MethodName} & \textbf{15.20} & \textbf{28.52} & \textbf{19.68} & \textbf{55.05} & \textbf{29.61\textcolor{green!60!black}{\scriptsize{\({\downarrow6.74}\)}}} \\

\addlinespace[0.3em]

GPT2-1.5B & 14.99 & 27.89 & 20.72 & 55.12 & 29.68 \\
GPT2-PM & 14.94 & 27.95 & 21.00 & 56.15 & 30.01 \\
ContextLM & 14.60 & 27.05 & 17.46 & \textbf{44.62} & \textbf{25.93} \\
\rowcolor[HTML]{F2F3F5} \textbf{\MethodName} & \textbf{14.40} & \textbf{26.80} & \textbf{17.26} & 47.06 & 26.38\textcolor{green!60!black}{\scriptsize{\({\downarrow3.30}\)}} \\

\midrule
\midrule

Pythia-70M & 18.27 & 57.01 & 142.01 & 973.59 & 297.72 \\
ContextLM & \textbf{14.96} & 43.64 & 71.45 & 440.77 & 142.71 \\
\rowcolor[HTML]{F2F3F5} \textbf{\MethodName} & 15.00 & \textbf{43.29} & \textbf{58.09} & \textbf{434.44} & \textbf{137.70\textcolor{green!60!black}{\scriptsize{\({\downarrow160.02}\)}}} \\

\addlinespace[0.5em]

Pythia-160M & 12.56 & 33.44 & 38.20 & 187.28 & 67.87 \\
ContextLM & 11.14 & 28.18 & 25.97 & 107.05 & 43.09 \\
\rowcolor[HTML]{F2F3F5} \textbf{\MethodName} & \textbf{10.93} & \textbf{27.01} & \textbf{20.01} & \textbf{79.32} & \textbf{34.23\textcolor{green!60!black}{\scriptsize{\({\downarrow33.54}\)}}} \\

\addlinespace[0.5em]

Pythia-410M & 8.88 & 20.11 & 10.85 & 31.53 & 17.84 \\
ContextLM & \textbf{8.67} & 19.50 & 10.15 & 24.95 & 15.82 \\
\rowcolor[HTML]{F2F3F5} \textbf{\MethodName} & 8.70 & \textbf{19.49} & \textbf{9.84} & \textbf{23.20} & \textbf{15.31\textcolor{green!60!black}{\scriptsize{\({\downarrow2.53}\)}}} \\

\bottomrule
\end{tabular}
\label{tab:GPT2_Pythia_PPL_res}
\end{table}

\begin{table*}[t]
    \centering
    \small
    \setlength{\tabcolsep}{4.5pt} 
    \caption{Evaluation of models across downstream tasks (Accuracy $\uparrow$) and language modeling (PPL $\downarrow$). \MethodName-GPT is denoted as \MethodName.}
    \renewcommand{\arraystretch}{1.2}
    \begin{tabular}{lccccccccc|c}
    \toprule
    \textbf{Model} & \makecell{Lambada \\ OpenAI} & \makecell{Lambada \\ Std.} & \makecell{ARC-\\E} & \makecell{ARC-\\C} & \makecell{Wino\\Gd.} & \makecell{PIQA} & \makecell{Hella-\\Swag} & \makecell{SciQ} & \makecell{RACE} & \textbf{Avg Acc} $\uparrow$ \\
    \midrule

    GPT-124M & 27.3 & 20.7 & 41.6 & 17.9 & 49.6 & 60.4 & 27.5 & 69.5 & 27.4 & 38.0 \\
    GPT-PM & 28.3 & 20.7 & 41.7 & 18.8 & 49.7 & 60.7 & 27.4 & 69.8 & 28.3 & 38.4 \\
    ConceptLM & 29.2 & 20.3 & 43.8 & 19.3 & 52.9 & 60.7 & 27.6 & 69.5 & 28.4 & \textbf{39.1} \\
    \rowcolor[HTML]{F2F3F5} \textbf{\MethodName} & 29.5 & 21.3 & 43.5 & 19.2 & 49.4 & 61.3 & 27.9 & 69.6 & 28.9 & 38.9\textcolor{green!60!black}{\scriptsize{${\uparrow0.9}$}}\\
    \addlinespace[0.3em]
    GPT-355M & 32.8 & 24.4 & 44.7 & 19.9 & 50.7 & 62.7 & 28.8 & 72.5 & 28.8 & 40.6 \\
    GPT-PM & 35.2 & 25.3 & 44.6 & 20.4 & 52.6 & 62.4 & 29.1 & 73.1 & 29.8 & 41.4 \\
    ContextLM & 36.8 & 26.7 & 46.7 & 19.5 & 50.9 & 62.9 & 29.8 & 74.2 & 29.3 & 41.9 \\
    \rowcolor[HTML]{F2F3F5} \textbf{\MethodName} & 36.6 & 27.1 & 46.0 & 20.7 & 50.6 & 64.2 & 30.3 & 75.0 & 29.1 & \textbf{42.3\textcolor{green!60!black}{\scriptsize{${\uparrow1.7}$}}}\\
    \addlinespace[0.3em]
    GPT-774M & 37.2 & 27.4 & 47.0 & 20.2 & 53.6 & 64.6 & 30.3 & 74.9 & 29.8 & 42.8 \\
    GPT-PM & 37.2 & 29.4 & 48.4 & 19.5 & 52.4 & 63.8 & 30.6 & 74.3 & 29.8 & 42.8 \\
    ContextLM & 40.7 & 29.9 & 47.9 & 21.1 & 51.3 & 62.9 & 29.8 & 76.6 & 31.0 & 43.9 \\
    
    \rowcolor[HTML]{F2F3F5} \textbf{\MethodName} & 41.1 & 30.6 & 48.5 & 19.9 & 52.1 & 65.3 & 31.4 & 77.0 & 30.0 & \textbf{44.0\textcolor{green!60!black}{\scriptsize{${\uparrow1.2}$}}}\\
    \addlinespace[0.3em]
    GPT-1.5B & 39.3 & 29.8 & 49.6 & 20.6 & 51.8 & 65.1 & 31.5 & 77.0 & 30.4 & 43.9 \\
    GPT-PM & 38.7 & 29.5 & 49.2 & 20.0 & 51.2 & 64.8 & 31.5 & 77.0 & 29.9 & 43.5 \\
    ContextLM & 41.9 & 32.3 & 49.9 & 21.4 & 51.9 & 66.1 & 32.3 & 77.7 & 31.2 & 45.0 \\
    \rowcolor[HTML]{F2F3F5} \textbf{\MethodName} & 42.9 & 32.2 & 49.9 & 21.9 & 52.1 & 66.8 & 32.4 & 76.8 & 30.6 & \textbf{45.1\textcolor{green!60!black}{\scriptsize{${\uparrow1.2}$}}}\\

    \midrule
    \midrule
    
    Pythia-70M & 18.3 & 13.4 & 36.9 & 18.5& 52.1 & 60.0& 26.6 & 60.5& 24.9 & 34.6\\
    ContextLM & 28.0 & 17.2 & 40.7 & 18.6 & 52.3 & 60.5 & 27.5 & 72.1 & 27.0 & \textbf{38.2}\\
    \rowcolor[HTML]{F2F3F5}
    \textbf{\MethodName} &  29.3 & 17.5  & 40.2 & 17.8 & 50.8 & 59.8 & 27.5 & 74.3 & 26.2 & \textbf{38.2\textcolor{green!60!black}{\scriptsize{${\uparrow3.6}$}}} \\
    
    \addlinespace[0.3em]
    Pythia-160M & 32.7 & 21.5 & 43.8 & 19.5 & 53.4 & 61.5 & 28.5 & 74.3 & 27.9 & 40.3 \\
    ContextLM & 37.2 & 25.6 & 45.0 & 19.5 & 52.0 & 62.9 & 29.2 & 77.6 & 28.7 & 42.0 \\
    \rowcolor[HTML]{F2F3F5}
    \textbf{\MethodName} & 41.8 & 25.9 & 44.8 & 18.3 & 53.4 & 62.1 & 29.4 & 80.0 & 30.1 & \textbf{42.9\textcolor{green!60!black}{\scriptsize{${\uparrow2.6}$}}} \\

    \addlinespace[0.3em]
    Pythia-410M & 51.6 & 36.4 & 52.2 & 21.3 & 53.9 & 66.8 & 33.8 & 81.2 & 30.7 & 47.5 \\
    ContextLM & 52.2 & 39.0 & 51.7 & 22.7 & 52.3 & 67.4 & 34.3 & 83.0 & 30.8 & 48.2 \\
    \rowcolor[HTML]{F2F3F5}
    \textbf{\MethodName} & 53.0 & 39.7 & 52.0 & 22.1 & 54.2 & 66.6 & 34.3 & 82.2 & 30.2 & \textbf{48.3\textcolor{green!60!black}{\scriptsize{${\uparrow0.8}$}}} \\


    \bottomrule
    \end{tabular}
    \label{tab:GPT2_Pythia_ACC_res}
\end{table*}

\textbf{GPT-2 Scaling Experiments:} For the GPT-2 series, we train three configurations from scratch for comparison: 
(i) Baseline: The standard GPT-2;
(ii) \MethodName: Our method;
(iii) Parameter-Matched (PM): The baseline with an additional two token-level layers to match the parameter count of \MethodName.  
Since we use very small codebooks, which have fewer than 2M parameters in total in the largest XL model, we do not match these parameters. We evaluate these models in scaling metrics, which we visualize in Figure~\ref{fig:scaling_triple}. Compared to the base models, 
we achieve the same performance with the largest base models using only 76\% of the tokens or 63\% of the parameters. And our model gains the improvement with significantly lower computational cost, requiring only 81\% used by the standard models.

The detailed results of language modeling tasks are shown at the top of Table~\ref{tab:GPT2_Pythia_PPL_res}, and the downstream tasks' performance is shown at the top of Table~\ref{tab:GPT2_Pythia_ACC_res}. We find that in language modeling tasks, our method achieves the best results except when testing Lambada-Standard on the XL-model. In downstream tasks, our model also consistently demonstrated the best average performance.

\textbf{Pythia Scaling Experiments:} For the Pythia series, we train \MethodName\ from scratch. Results for the baseline and our strong baseline ContextLM, a concept-level model without explicit but implicit next concept prediction, are sourced from~\citet{dai2025contextlevellanguagemodelinglearning}. To ensure a fair comparison, we align all training and evaluation settings. The language modeling performance of Pythia series is reported at the bottom of Table~\ref{tab:GPT2_Pythia_PPL_res}, and the downstream performance is shown at the bottom of Table~\ref{tab:GPT2_Pythia_ACC_res}.

We find that the average performance on downstream tasks of our model exceeds the baselines. For language modeling tasks, our model also generally outperforms the base model. While the 70M and 410M models perform slightly worse than ContextLM on the trained Pile dataset, our method demonstrates superior generalization in language modeling, outperforming ContextLM across several other datasets.

We suggest that the generalization performance of ConceptLM is attributed to the introduction of the concept vocabulary, which saves fundamental conceptual representations in a stable, discrete space after massive data training, thereby exhibiting significant generalization. 
Notably, the use of a fixed configuration ($S$ and $N$) across all the models may not be optimal for every model. Our 160M model achieves a more pronounced improvement over the baseline compared to other scales, which we attribute to a potentially optimal alignment with the settings. 
This further underlines the robustness and potential of our proposed method.

\subsection{Continual Pre-train ConceptLM on 8B models}
\label{sub_sec:8b_models}

To investigate the scalability and generalizability of NCP, we extend the Llama-3.1-8B with concept-level prediction through continual pre-training. The model is trained on 9.6B tokens. We compare three variants: (i) Trained: The original Llama-3.1-8B, further trained on our corpus. (ii) \MethodName: The model augmented with two concept-level layers and the concept vocabulary, we also use an $S=\mathrm{the \ head \ numbers}$ and $N=64$ setting. 
(iii) PM: The parameter-matching token-level model. To ensure stable learning during the expansion, we initialize the newly added layers using the weights of the first two layers of the base model. We evaluate these models across widely used benchmarks, including MMLU~\citep{hendrycks2021measuringmassivemultitasklanguage}, ARC-Challenge, AGIEval~\citep{zhong2023agievalhumancentricbenchmarkevaluating}, and SQuAD 2.0~\citep{rajpurkar2018knowdontknowunanswerable}.

We show the 8B-parameter experiments in Table~\ref{tab:conceptlm_Llama_results}, \MethodName\ gains a 0.4 average improvement compared to the strong baseline PM model, and gains a 0.1 average improvement compared to the Trained-8B model. It is worth noting that training new layers within a pre-trained LLM is inherently challenging due to the difficulty of aligning new parameters with well-established representations. Despite these difficulties, our results demonstrate that concept-level prediction further enhances the performance of existing token-level models, even at an 8B-parameter scale with only 9.6B tokens.
In addition, we observe that the $\mathcal{L}_\text{VQ}$ continues to decrease with training, indicating a better fit to the continuous concept space. This suggests that combining explicit concept-level prediction with token-level models may yield more significant gains when scaled to larger datasets.

\section{Ablation Studies and Analysis}

\begin{table}[htbp]
    \centering
    \setlength{\tabcolsep}{2.2pt}  
    \caption{Results of continual pre-trained Llama-3.1-8B with 9.6B tokens. We denote \MethodName as Ours. We train the models using the same settings. Trained-Base-8B denotes the continual pretrained base model; PM-8B denotes the token-level base model with 2 more token-level layers to match the parameter with \MethodName.}
    \vspace{0.5em}
    \renewcommand{\arraystretch}{1.3}
    \begin{tabular}{lccccc}
    \toprule
    Model & MMLU & ARC-C & AGIEval & SquAD 2.0 & Avg \\
    \midrule
    Trained-8B & 64.3 & 34.8 & \textbf{54.2} & \textbf{36.7} & 47.5 \\
    PM-8B & 64.5 & 35.0 & 53.3 & 36.1 & 47.2 \\
    \rowcolor[HTML]{F2F3F5}
    \textbf{Ours-8B} & \textbf{64.6\textcolor{green!60!black}{\scriptsize{\(\)}}} & \textbf{35.1} \textcolor{green!60!black}{\scriptsize{\(\)}} & \textbf{54.2\textcolor{green!60!black}{\scriptsize{\(\)}}} & 36.5\textcolor{green!60!black}{\scriptsize{\(\)}}
    & \textbf{47.6\textcolor{green!60!black}{\scriptsize{\({}\)}}} \\
    \bottomrule
    \end{tabular}
    \label{tab:conceptlm_Llama_results}
\end{table}


We conduct ablation studies and analysis in this Section. For the ablation studies, we discuss the structural and functional design of our model, including: (i) the concept-level modeling and the concept vocabulary, and (ii) the training objectives. 
For the analysis, we analyze the features of our method. Specifically, we analyze its performance in long-range modeling and its scaling performance compared to standard token-level models. Finally, we provide an assessment of the computational costs of our method.
Unless otherwise specified, we use a 160M Pythia model as the backbone and train it from scratch on 30B tokens.

\subsection{Concept-level Modeling and Concept Vocabulary}

To evaluate the effectiveness of each component, we compare three configurations under the same experimental settings, as reported in Table~\ref{tab:GPT2_Pythia_PPL_res} and Table~\ref{tab:GPT2_Pythia_ACC_res}. Specifically, the comparison between the PM (Parameter-Matching) models and \MethodName\ models highlights the superiority of concept-level modeling in enhancing model capabilities. Furthermore, the performance gap between ContextLM and \MethodName\ on the Pythia experiments underscores the generalization advantages of explicitly incorporating a concept vocabulary compared to implicit concept modeling.

\begin{table}[ht]
    \centering
    \small
    \setlength{\tabcolsep}{4.5pt} 
    \caption{Loss ablation results. We present a training objective ablation study for ConceptLM and a baseline $\mathcal{L}_{\text{MTP}}$ result.}
    \resizebox{\textwidth}{!}{
    \renewcommand{\arraystretch}{1.2}
    \begin{tabular}{lccccccccc|c}
    \toprule
    \textbf{Model} & \makecell{Lambada \\ OpenAI} & \makecell{Lambada \\ Std.} & \makecell{ARC-\\E} & \makecell{ARC-\\C} & \makecell{Wino\\Gd.} & \makecell{PIQA} & \makecell{Hella-\\Swag} & \makecell{SciQ} & \makecell{RACE} & \textbf{Avg Acc} $\uparrow$ \\
    \midrule
    \multicolumn{11}{c}{\textit{Zero-shot Evaluation (Accuracy \%)}} \\
    \midrule
    
    \multicolumn{11}{c}{\textbf{\textit{Baseline}}} \\
    $\mathcal{L}_{\text{MTP}}$ & 31.9 & 20.4 & 40.2 & 17.7 & 49.3 & 60.3 & 27.9 & 71.2 & 27.0 & 38.4 \\
    \multicolumn{11}{c}{\textbf{\textit{ConceptLM}}} \\
    
    $\mathcal{L}_{\text{NTP}}$ & 32.9 & 20.7 & 43.4 & 18.0 & 50.7 & 61.3 & 28.0 & 72.3 & 27.0 & 39.6 \\

    $\mathcal{L}_{\text{NTP}}$ + $\mathcal{L}_{\text{VQ}}$ & 34.4 & 20.5 & 43.4 & 18.0 & 49.1 & 61.4 & 28.0 & 72.3 & 27.6 & 39.8 \\

    $\mathcal{L}_{\text{NTP}}$ + $\mathcal{L}_{\text{NCP}}$ & 34.0 & 20.5 & 43.1 & 18.3 & 49.4 & 61.3 & 28.1 & 73.2 & 27.9 & 39.5 \\

    \rowcolor[HTML]{F2F3F5} $\mathcal{L}_{\text{NTP}}$ + $\mathcal{L}_{\text{NCP}}$ +  $\mathcal{L}_{\text{VQ}}$ & 34.0 & 21.8 & 43.5 & 18.8 & 53.0 & 61.3 & 28.3 & 73.3 & 28.6 & \textbf{40.3} \\
    
    \midrule
    \midrule
    \textbf{Model} & \multicolumn{2}{c}{OWT} & \multicolumn{2}{c}{Wikitext} & \multicolumn{2}{c}{Lambada OpenAI} & \multicolumn{2}{c}{Lambada Std} & --- & \textbf{Avg PPL} $\downarrow$ \\
    \midrule
    \multicolumn{11}{c}{\textit{Language Modeling (Perplexity)}} \\
    \midrule

    \multicolumn{11}{c}{\textbf{\textit{Baseline}}} \\
    $\mathcal{L}_{\text{MTP}}$ & \multicolumn{2}{c}{14.83 } & \multicolumn{2}{c}{36.51} & \multicolumn{2}{c}{43.12} & \multicolumn{2}{c}{230.18} & --- & 81.16 \\
    
    \multicolumn{11}{c}{\textbf{\textit{ConceptLM}}} \\

    $\mathcal{L}_{\text{NTP}}$ & \multicolumn{2}{c}{12.64 } & \multicolumn{2}{c}{34.30} & \multicolumn{2}{c}{37.31} & \multicolumn{2}{c}{192.87} & --- & 69.28 \\

    $\mathcal{L}_{\text{NTP}}$ + $\mathcal{L}_{\text{VQ}}$ & \multicolumn{2}{c}{12.62 } & \multicolumn{2}{c}{33.66} & \multicolumn{2}{c}{37.60} & \multicolumn{2}{c}{218.69} & --- &  75.64 \\

    $\mathcal{L}_{\text{NTP}}$ + $\mathcal{L}_{\text{NCP}}$ & \multicolumn{2}{c}{13.21 } & \multicolumn{2}{c}{35.95} & \multicolumn{2}{c}{41.35} & \multicolumn{2}{c}{214.70} & --- & 76.30 \\

    \rowcolor[HTML]{F2F3F5} $\mathcal{L}_{\text{NTP}}$ + $\mathcal{L}_{\text{NCP}}$ +  $\mathcal{L}_{\text{VQ}}$ & \multicolumn{2}{c}{12.64 } & \multicolumn{2}{c}{33.45} & \multicolumn{2}{c}{36.38} & \multicolumn{2}{c}{190.06} & --- &  \textbf{68.13} \\

    \bottomrule
    \end{tabular}
    }
    \label{tab:loss_abl_detailed}
\end{table}

\subsection{Training Objectives}
\label{subsec:loss_abl}

We conduct an ablation study by systematically removing $\mathcal{L}_{\text{NCP}}$ and $\mathcal{L}_{\text{VQ}}$ to evaluate the contribution of each loss, while keeping the model architecture unchanged. As shown in Table~\ref{tab:loss_abl_detailed}, using $\mathcal{L}_{\text{NTP}}$ alone provides the model with solid foundational capabilities. The independent using of either $\mathcal{L}_{\text{VQ}}$ or $\mathcal{L}_{\text{NCP}}$ leads to a performance drop. This suggests that without quantization constraints, the vocabulary fails to learn robust concept representations; conversely, without concept-prediction supervision, the model cannot effectively predict useful discrete concepts. Optimal performance is only achieved when objectives are combined.

Furthermore, we train a token-level model with the MTP loss $\mathcal{L}_{\text{MTP}}$, $M=4$ to serve as one of our baselines; however, empirical evidence, including the original MTP paper, suggests that MTP models do not yield superior performance at scales below 3B parameters, larger than our largest 1.5B pretrained model. Consequently, rather than categorizing it as a strong baseline in the main experiments, we train a small MTP variant to evidence this in Table~\ref{tab:loss_abl_detailed}.

\subsection{Long-range Modeling Analysis} 
 
We assess long-context modeling capabilities by increasing training sequence lengths while keeping all 
\begin{wrapfigure}{r}{0.48\linewidth}
  \centering
  \vspace{-0.5em}  
  \includegraphics[width=\linewidth]{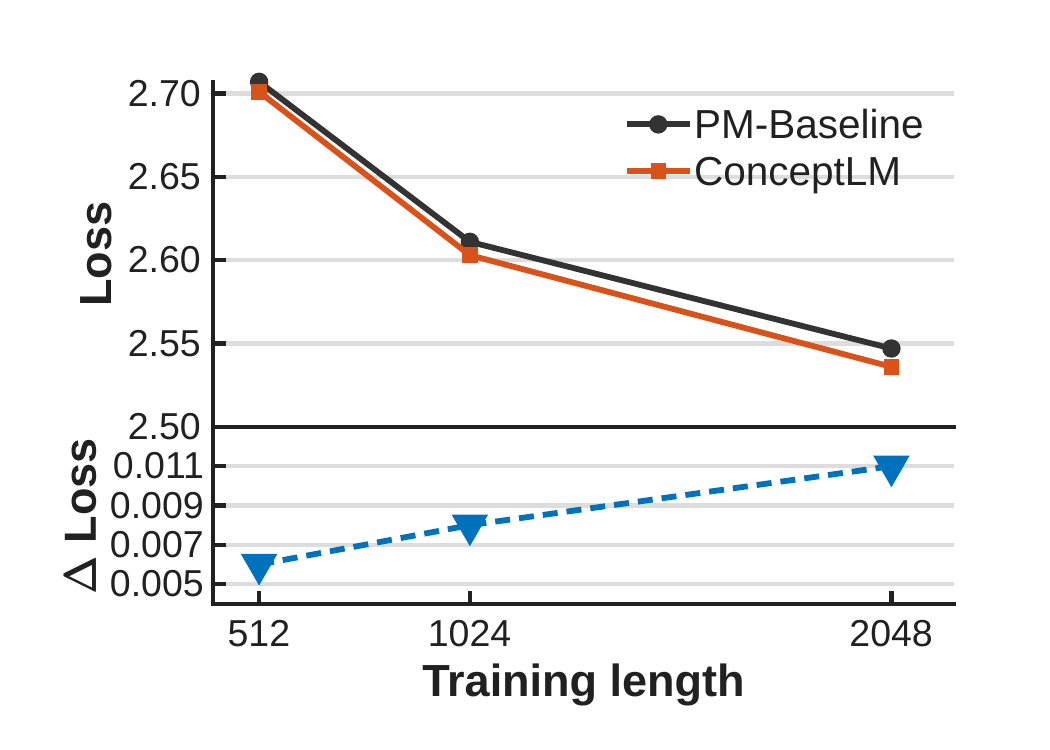}
  \caption{Long-range dependency experiments.}
  \label{fig:long_range_dependency}
  \vspace{-0.5em}  
\end{wrapfigure}
other settings fixed. We compare the language modeling loss of \MethodName\ and the PM baseline after training on token length from 512 to 2048.
As shown in Figure~\ref{fig:long_range_dependency}, \MethodName\ consistently achieves larger improvements over the PM-Baseline as the sequence length increases. 
This highlights a critical advantage of concept-level modeling beyond general performance gains: the concept-level module operates on a sequence length reduced by a factor of $k$. This reduction allows the computational overhead to be more effectively focused on capturing long-range dependencies rather than being focused on the local tokens.

\begin{table}[ht]
    \centering
    \small
    \setlength{\tabcolsep}{4.5pt} 
    \caption{Comparison of model scaling w/ or w/o the concept layers. }
    \renewcommand{\arraystretch}{1.2}
    \resizebox{\textwidth}{!}{
    \begin{tabular}{lccccccccc|c}
    \toprule
    \textbf{Model} & \makecell{Lambada \\ OpenAI} & \makecell{Lambada \\ Std.} & \makecell{ARC-\\E} & \makecell{ARC-\\C} & \makecell{Wino\\Gd.} & \makecell{PIQA} & \makecell{Hella-\\Swag} & \makecell{SciQ} & \makecell{RACE} & \textbf{Avg Acc} $\uparrow$ \\
    \midrule
    \multicolumn{11}{c}{\textit{Zero-shot Evaluation (Accuracy \%)}} \\
    \midrule
    
    +2 token-layers & 33.7 & 21.0 & 44.5 & 20.1 & 50.7 & 61.4 & 27.9 & 74.7 & 27.4 & 40.1 \\

    \rowcolor[HTML]{F2F3F5} +2 concept-layers & 34.0 & 21.8 & 43.5 & 18.8 & 53.0 & 61.3 & 28.3 & 73.3 & 28.6 & \textbf{40.3}\textcolor{green!60!black}{\scriptsize{\(+0.2\)}} \\

    +6 token-layers & 34.7 & 22.5 & 44.8 & 18.8 & 52.1 & 60.2 & 27.6 & 72.3 & 28.6 & 40.1 \\
    
    \rowcolor[HTML]{F2F3F5} +6 concept-layers & 33.9 & 23.3 & 43.0 & 19.7 & 54.1 & 61.0 & 28.4 & 76.0 & 28.6 & \textbf{40.9}\textcolor{green!60!black}{\scriptsize{\(+0.8\)}} \\

    +10 token-layers & 33.0 & 20.7 & 44.0 & 19.8 & 52.0 & 61.9 & 28.1 & 74.7 & 27.0 & 40.2 \\

    \rowcolor[HTML]{F2F3F5} +10 concept-layers & 33.9 & 22.3 & 43.3 & 18.7 & 53.6 & 61.5 & 28.3 & 73.8 & 27.9 & \textbf{40.4}\textcolor{green!60!black}{\scriptsize{\(+0.2\)}} \\

    \midrule
    \midrule
    \textbf{Model} & \multicolumn{2}{c}{OWT} & \multicolumn{2}{c}{Wikitext} & \multicolumn{2}{c}{Lambada OpenAI} & \multicolumn{2}{c}{Lambada Std} & --- & \textbf{Avg PPL} $\downarrow$ \\
    \midrule
    \multicolumn{11}{c}{\textit{Language Modeling (Perplexity)}} \\
    \midrule

    +2 token-layers & \multicolumn{2}{c}{12.77 } & \multicolumn{2}{c}{34.02} & \multicolumn{2}{c}{37.94} & \multicolumn{2}{c}{200.67} & --- &  59.80 \\

    \rowcolor[HTML]{F2F3F5} +2 concept-layers & \multicolumn{2}{c}{12.64 } & \multicolumn{2}{c}{33.45} & \multicolumn{2}{c}{36.38} & \multicolumn{2}{c}{190.06} & --- &  \textbf{68.13\textcolor{green!60!black}{\scriptsize{\(+3.22\)}}} \\

    +6 token-layers & \multicolumn{2}{c}{12.74 } & \multicolumn{2}{c}{34.06} & \multicolumn{2}{c}{37.31} & \multicolumn{2}{c}{176.90} & --- &  65.25 \\

    \rowcolor[HTML]{F2F3F5} +6 concept-layers & \multicolumn{2}{c}{12.45 } & \multicolumn{2}{c}{33.06} & \multicolumn{2}{c}{37.83} & \multicolumn{2}{c}{134.48} & --- &  \textbf{54.34\textcolor{green!60!black}{\scriptsize{\(+7.91\)}}} \\

    +10 token-layers & \multicolumn{2}{c}{12.73 } & \multicolumn{2}{c}{34.06} & \multicolumn{2}{c}{40.69} & \multicolumn{2}{c}{213.06} & --- &  75.20 \\

    \rowcolor[HTML]{F2F3F5} +10 concept-layers & \multicolumn{2}{c}{12.32 } & \multicolumn{2}{c}{33.26} & \multicolumn{2}{c}{36.96} & \multicolumn{2}{c}{150.88} & --- &  \textbf{58.21\textcolor{green!60!black}{\scriptsize{\(+16.99\)}}} \\

    \bottomrule
    \end{tabular}
    }
    \label{tab:scaling_ana_detailed}
\end{table}

\subsection{Scaling Performance Analysis} 
\label{subsec:scaling}

We show that introducing concept-level modeling significantly enhances scaling behavior. As shown in Table~\ref{tab:scaling_ana_detailed}, we observe that for standard token-level models, simply stacking additional token layers does not yield stable performance gains. The performance even degrades significantly when the depth reaches 22 token-level layers. This phenomenon can be attributed to the depth-width imbalance or representation bottlenecks, where the limited dimension cannot effectively support the increased abstraction required by excessive depth, leading to worse performance.

However, our results demonstrate that by incorporating concept-level modeling with a certain number of token layers, \MethodName\ effectively overcomes these scaling limitations. With the concept-level layers operating on a denser semantic space, \MethodName\ exhibits a widening performance PPL.Avg gap as depth increases, underscoring its robust scaling behavior under a fixed-width configuration.

\subsection{Computational Complexity Analysis}
\label{subsec:efficiency_analysis}

We analyze the computational complexity of \MethodName\ in Table~\ref{tab:llama3_train_infer}. The empirical training cost shows our training time of the 8B models with 8 NVIDIA H200 GPUs, and the inference cost is the time test on the SQuAD-V2 benchmark. Theoretically, by compressing the sequence length by a factor of $k$, the concept-level layers achieve a quadratic reduction ($k^2$) in attention overhead and a linear reduction ($k$) in MLP computation. Notably, this advantage becomes more substantial as the sequence length $T$ increases.

\begin{table}[htbp]
    \centering
    \caption{Training and Inference cost comparison. $T$ denotes sequence length, $d$ denotes the hidden dimension.} 
    \label{tab:llama3_train_infer}
    \renewcommand{\arraystretch}{1.3}
    \setlength{\tabcolsep}{1pt}
    \begin{tabular}{lcc}
    \toprule
    \textbf{Layer / Model} & \textbf{Training Cost} & \textbf{Inference Cost} \\
    \midrule
    \rowcolor[HTML]{F9F9F9} \multicolumn{3}{l}{\textit{Theoretical Complexity Analysis}} \\
    Token Layer    & $O(T^2d + Td^2)$ & $O(Td + d^2)$ \\

    \rowcolor[HTML]{F2F3F5}
    Concept Layer & $O((\frac{T}{k})^2d + \frac{T}{k}d^2)$ & $O(\frac{T}{k}d + d^2)$ \\
    \midrule
    \rowcolor[HTML]{F9F9F9} \multicolumn{3}{l}{\textit{Empirical Cost (Llama-3.1-8B)}} \\
    Llama-PM-8B    & 80.0 h & 99.0 min \\
    
    \rowcolor[HTML]{F2F3F5}
    \textbf{\MethodName-8B} & \textbf{71.0 h\textcolor{green!60!black}{\scriptsize{${\downarrow11.3\%}$}}} & \textbf{86.0 min\textcolor{green!60!black}{\scriptsize{${\downarrow13.1\%}$}}} \\
    \bottomrule
    \end{tabular}
\end{table}

Practically, in our experiment, the Llama-PM-8B model is a 34-layer token-level model, \MethodName\ replacing 2 out of the 34 token-level layers with 2 concept-level layers, which should yield a training computational reduction of approximately 4.69\% in terms of FLOPs, except for the embedding in/out module. However, our empirical results show a more significant improvement (11.3\% in training and 13.1\% in inference). This is attributed to the substantial reduction in IO cost. By processing concept representations instead of token representations, \MethodName\ reduces the KV cache size and intermediate hidden states size in the concept layers, thereby alleviating the memory bandwidth bottleneck that often dominates large-scale model training. We provide the FLOPs calculation in Appendix~\ref{App:Comp_cost}.

\section{Related Work}

\subsection{Abstract-level Prediction in AI}

A growing body of research explores high-level prediction in AI, shifting learning objectives from fine-grained, low-level targets to abstract semantic representations. As a pioneering framework, JEPA~\citep{assran2023selfsupervisedlearningimagesjointembedding, bardes2024revisitingfeaturepredictionlearning} emphasizes prediction within a joint-embedding space to bypass the modeling of redundant pixel-level details, a paradigm successfully extended to video understanding~\citep{assran2025vjepa2selfsupervisedvideo}. In the realm of decision-making and planning, world models such as Dreamer~\citep{hafner2024masteringdiversedomainsworld} and GAIA-1~\citep{hu2023gaia1generativeworldmodel} demonstrate that predicting future latent states, rather than raw inputs, is crucial for robust long-horizon reasoning. Similarly, the field of audio has transitioned from raw waveform reconstruction toward the prediction of discrete hidden units~\citep{baevski2020wav2vec20frameworkselfsupervised, hsu2021hubertselfsupervisedspeechrepresentation} to better capture long-range dependencies. And for multimodal learning, the Large Concept Model (LCM)~\citep{lcmteam2024largeconceptmodelslanguage} operates in a modality-agnostic latent space, performing inference on sentence-level embeddings rather than tokens. Collectively, these developments suggest that higher level prediction is a universal path toward building better models.

\subsection{Abstract-level Modeling in Language Models}

There are also several works that have explored language modeling beyond individual tokens. One prominent approach involves structural compression, where tokens are aggregated into patches: MegaByte~\citep{yu2023megabytepredictingmillionbytesequences} uses a fixed chunk size to convert bytes into patches; BLT~\citep{pagnoni2024bytelatenttransformerpatches} and H-net~\citep{hwang2025dynamicchunkingendtoendhierarchical} dynamically patchify byte sequences into patches, and DLCM~\citep{qu2026dynamiclargeconceptmodels} extends the H-net paradigm to the token level.
They then perform patch-level language modeling. 
The high-level process methods also focus on accelerating model training or inference~\citep{shao2025tokenpredictionpatchleveltraining, liu2025hamburgeracceleratingllminference}. And more recently, latent reasoning paradigms~\citep{su2025tokenassortedmixinglatent, hao2025traininglargelanguagemodels, tack2025llmpretrainingcontinuousconcepts, deng2024explicitcotimplicitcot} emerged to leverage latent tokens to capture abstract representations. Despite these advancements, designing a self-supervised pretraining objective specifically for high-level language representations remains an open challenge.

\section{Conclusion}

In this paper, we propose Next Concept Prediction (NCP), a novel and more challenging pre-training objective built upon a discrete concept-level latent space. By leveraging VQ to construct this space and integrating it with standard Next Token Prediction (NTP), we develop the \MethodName\ framework. 
Our results demonstrate that \MethodName\ exhibits superior performance on language modeling tasks and downstream tasks than standard token-level models and strong baselines.
Moreover, by leveraging continual pre-training to transform an 8B existing token-level model into a context-aware variant, we show the applicability, robustness, and low-cost feature of our method.
Finally, we provide empirical evidence of the model’s long-range modeling ability and its scalability across different parameter scales.
These findings suggest that the modeling supervision space transitioning, from token-level to concept-level, is a promising path for building the next generation of LLMs.

\section*{Acknowledgment}
This work is sponsored by the National Natural Science Foundation of China (NSFC) grant (No. 62576211) and the National Key Research and Development Program of China (No. 2023ZD0121402). It is also the result of a collaborative project on novel language model architectures between Shanghai Jiao Tong University (SJTU) and the Shanghai Artificial Intelligence Laboratory. The computational resources required for pretraining the models were provided by the Shanghai AI Lab.

\clearpage
\bibliography{main}

\newpage

\appendix
\onecolumn          

\section{NCP versus NTP}

We show the comparison of NCP versus NTP in Table~\ref{tab:ntp_vs_ncp}.

\begin{table}[htbp]
    \centering
    \caption{NTP v.s. NCP}
    \label{tab:ntp_vs_ncp}
    \renewcommand{\arraystretch}{1.3} 
    \small
    \resizebox{\textwidth}{!}{
    \begin{tabular}{lcc}
        \toprule
        \textbf{Module} & \textbf{Next Token Prediction LM} & \textbf{Next Concept Prediction LM} \\
        \midrule
        
        Input & Discrete tokens & Token Hidden states \\
        
        Initial Representation & Standard token embedding & Concept embedding grouped by hidden states \\
        
        Backbone & Transformers & Transformers \\
        
        LM Head & Token hidden states $\rightarrow{}$ token logits & Concept hidden states $\rightarrow{}$ concept logits \\

        Vocabulary & Pre-trained BPE vocabulary & Jointly optimized concept vocabulary \\

        Generation Strategy & Sampling from vocabulary by logits  & Aggregating over vocabulary by logits  \\
        
        Loss Function & Cross-Entropy Loss & Mean Squared Error Loss \\
        \bottomrule
    \end{tabular}
    }
\end{table}

\section{Experiments Setting}
\label{App:exp_setting}

\subsection{Model Setup}

We list the model backbones, training stages, training datasets, and tokens in Table~\ref{tab:exp_summary}.

\begin{table}[htbp]
    \centering
    \caption{Experiments Setting Summary.}
    \label{tab:exp_summary}
    \begin{tabular}{c c c c c}
        \toprule
        Model Backbone & Parameters & Training Stage & Training Dataset & Token Numbers \\
        \midrule
        GPT-2  & 160M-1.5B & Pre-training          & OpenWebText & 8B    \\
        Pythia & 70M-410M & Pre-training           & The Pile    & 300B  \\
        
        Llama  & 8B     & Continual Pre-training& LongCtxEng  & 9.6B  \\
        \bottomrule
    \end{tabular}
\end{table}

\subsection{ConceptLM Setup}
\label{app:conceptlm_setup}

We list the specific setting for ConceptLM in Table~\ref{tab:clm_summary}, including insert depth, concept layers, chunk size, $S$, and $N$. 

\paragraph{Insert Depth Discussion: } To ensure that more Decoder layers benefit from multi-level information while maintaining a token-level feature mixing before the concept layer, we default to inserting concept layers after the first Token Layer of the backbone. In this configuration, the first layer acts as the Token-level Encoder, while the remaining $L-1$ layers serve as Token-level Decoder.

However, given that the GPT-2 model was pretrained on a relatively small dataset (less than 10B tokens), its performance is highly sensitive to architectural depth. To ensure a fair comparison with our strong baseline, ContextLM, we conducted additional experiments on GPT-2 by inserting the concept layer before the first Token-level layer. Results for the configuration where the concept layer is inserted after the first layer are reported in Appendix~\ref{app:GPT2_res_L1}.

Furthermore, we conducted extensive ablation studies to analyze the impact of the insert depth. Our results indicate that placing the concept layers at either the very beginning (Layer 0) or after the first layer (Layer 1) is suboptimal. Instead, the model achieves the best performance when the layer is positioned at three-quarters of the backbone depth (a trend consistently verified through repeat experiments). We hypothesize that this superiority stems from the backpropagation of concept-level supervision into the Encoder layers; a sufficiently deep Encoder provides more refined representations, thereby facilitating more accurate concept prediction. Detailed experimental results are provided in Appendix~\ref{app:insert_depth_abl}.

\begin{table*}[htbp]
    \centering
    \caption{ConceptLM Setting Summary in main experiments.}
    \label{tab:clm_summary}
    \resizebox{\textwidth}{!}{
    \begin{tabular}{c c c c c c}
        \toprule
        Model Backbone & Insert Depth & Concept Layers & Chunk Size & $S$ & $N$ \\
        \midrule
        GPT-2  & before the first/second layer & 2 & 4 & head number  & 64\\
        Pythia & before the second layer & 2 & 4 & head number & 64 \\
        
        Llama  & before the second layer & 2 & 4 & head number & 64 \\
        \bottomrule
    \end{tabular}
    }
\end{table*}


\subsection{Training Configuration for GPT2 Backbone}
\label{App:GPT_conf}

We show the training configuration of GPT-2 series models, all the variant of a certain size (base model, PM model, \MethodName, ContextLM) use the same training setting in Table~\ref{tab:GPT2_config}.

\begin{table*}[htbp]
    \centering
    \caption{GPT-2 training configuration.}
    \label{tab:GPT2_config}
    \begin{tabular}{c c c c}
        \toprule
        Model Size & Learning Rate & Batch Size & Training Length \\
        \midrule
        GPT-Base (124M)  & 1e-3 & 512 & 1024 \\
        GPT-Medium (355M) & 8e-4 & 512 & 1024 \\
        GPT-Large (774M) & 6e-4 & 512 & 1024 \\
        GPT-XL (1.5B) & 4e-4 & 512 & 1024 \\
        \bottomrule
    \end{tabular}
\end{table*}

\subsection{Training Configuration for Pythia Backbone}
\label{App:Pythia_conf}

We show the training configuration of Pythia series models, all the variant of a certain size (base model, PM model, \MethodName, ContextLM) use the same training setting in Table~\ref{tab:Pythia_config}.

\begin{table*}[htbp]
    \centering
    \caption{Pythia training configuration.}
    \label{tab:Pythia_config}
    \begin{tabular}{c c c c}
        \toprule
        Model Size & Learning Rate & Batch Size & Training Length \\
        \midrule
        Pythia-70M   & 1e-3 & 1024 & 2048 \\
        Pythia-160M  & 6e-4 & 1024 & 2048 \\
        Pythia-410M  & 4e-4 & 1024 & 2048 \\
        \bottomrule
    \end{tabular}
\end{table*}

\subsection{Training Configuration for Llama Model}
\label{App:Llama_conf}

We show the training configuration of Llama model, all the variant of a certain size (base model, PM model, \MethodName) use the same training setting.

\begin{table*}[htbp]
    \centering
    \caption{Llama training configuration.}
    \label{tab:Llama_config}
    \begin{tabular}{c c c c}
        \toprule
        Model Size & Learning Rate & Batch Size & Training Length \\
        \midrule
        Llama-3.1-8B   & 1e-5 & 96 & 8192 \\
        \bottomrule
    \end{tabular}
\end{table*}

\section{8B Models Computation Cost}
\label{App:Comp_cost}

We provide a brief computation cost analysis, mainly use the following function, where $T$ is the input sequence length, $d$ is the hidden size.

$$\begin{aligned}
\text{Token-level Layer FLOPs} &= \underbrace{8Td^2}_{\text{Attention Linear Layers}} + \underbrace{4Td^2}_{\text{FFN (at } 4d)} + \underbrace{2T^2d}_{\text{Attention Score}} \\
&= 12Td^2 + 2T^2d
\end{aligned}$$

For a certain 34 token-layers Llama-3.1-8B-PM, and a 32 token-layers and 2 concept-layers ConceptLM-Llama-3.1-8B.

Llama-3.1-8B-PM:

$$\text{Total FLOPs}_{PM} = 34 \times (12Td^2 + 2T^2d) = 408Td^2 + 68T^2d$$

ConceptLM-Llama-3.1-8B:

$$\begin{aligned}
\text{Total FLOPs}_{ConceptLM} &= 32 \times (12Td^2 + 2T^2d) + 2 \times (3Td^2 + \frac{1}{8}T^2d) \\
&= (384Td^2 + 64T^2d) + (6Td^2 + \frac{1}{4}T^2d) \\
&= 390Td^2 + 64.25T^2d
\end{aligned}$$

With our training setting $T=8192, h=4096$, we have

$$\text{Ratio} = \frac{\text{Total FLOPs}_{PM} -\text{Total FLOPs}_{ConceptLM}}{\text{Total FLOPs}_{PM}} * 100\% \approx 4.69\%$$

\section{Downstream Tasks}
\label{App:downstream_tasks}

We show the evaluation capabilities for benchmarks.

\begin{table}[h]
\centering
\caption{Evaluated benchmarks with capabilities.}
\resizebox{\textwidth}{!}{
\begin{tabular}{l l}
\toprule
\textbf{Capability} & \textbf{Benchmarks / Datasets} \\
\midrule
Long-range dependency \& context consistency & Lambada (OpenAI / Standard) \\
Commonsense reasoning & ARC-Easy, ARC-Challenge, WinoGrande, PIQA, HellaSwag \\
Scientific knowledge & SciQ \\
Paragraph comprehension & RACE \\
General language modeling & The Pile, WikiText, OpenWebText \\
\bottomrule
\end{tabular}
}
\label{tab:eval_capabilities}
\end{table}

\section{Experimental Results}

We show other experimental results in this section.

\subsection{GPT2 Results with L1 setting}
\label{app:GPT2_res_L1}

We show all the GPT-2 experiments, including insert concept-level layers after the first layer in Table~\ref{tab:conceptlm_GPT_results_L1}.

\begin{table}[t]
    \centering
    \small
    \setlength{\tabcolsep}{1pt} 
    \caption{Evaluation of models across downstream tasks (Accuracy $\uparrow$) and language modeling (PPL $\downarrow$). \MethodName-GPT is denoted as \MethodName. \MethodName-L1 denotes that we insert the concept-level module after the first layer.
    \MethodName denotes that we insert the concept-level module before the first layer.}
    \resizebox{\textwidth}{!}{
    \begin{tabular}{lccccccccc|l}
    \toprule
    \textbf{Model} & \makecell{Lambada \\ OpenAI} & \makecell{Lambada \\ Std.} & \makecell{ARC-E} & \makecell{ARC-C} & \makecell{Wino\\Gd.} & \makecell{PIQA} & \makecell{Hella-\\Swag} & \makecell{SciQ} & \makecell{RACE} & \textbf{Avg Acc} $\uparrow$ \\
    \midrule
    \multicolumn{11}{c}{\textit{Zero-shot Evaluation (Accuracy \%)}} \\
    \midrule
    GPT-Base & 27.3 & 20.7 & 41.6 & 17.9 & 49.6 & 60.4 & 27.5 & 69.5 & 27.4 & 38.0 \\
    GPT-Base-PM & 28.3 & 20.7 & 41.7 & 18.8 & 49.7 & 60.7 & 27.4 & 69.8 & 28.3 & 38.4 \\
    ConceptLM-Base & 29.2 & 20.3 & 43.8 & 19.3 & 52.9 & 60.7 & 27.6 & 69.5 & 28.4 & \textbf{39.1} \\
    \rowcolor[HTML]{F2F3F5} \textbf{\MethodName-Base-L1} & 28.9 & 20.7 & 41.9 & 18.0 & 50.0 & 60.2 & 27.8 & 70.8 & 30.1 & 38.7\textcolor{green!60!black}{\scriptsize{${+0.7}$}}\\
    \rowcolor[HTML]{F2F3F5} \textbf{\MethodName-Base} & 29.5 & 21.3 & 43.5 & 19.2 & 49.4 & 61.3 & 27.9 & 69.6 & 28.9 & 38.9\textcolor{green!60!black}{\scriptsize{${+0.9}$}}\\
    \addlinespace[0.3em]
    GPT-medium & 32.8 & 24.4 & 44.7 & 19.9 & 50.7 & 62.7 & 28.8 & 72.5 & 28.8 & 40.6 \\
    GPT-medium-PM & 35.2 & 25.3 & 44.6 & 20.4 & 52.6 & 62.4 & 29.1 & 73.1 & 29.8 & 41.4 \\
    ContextLM-medium & 36.8 & 26.7 & 46.7 & 19.5 & 50.9 & 62.9 & 29.8 & 74.2 & 29.3 & 41.9 \\
    \rowcolor[HTML]{F2F3F5} \textbf{\MethodName-Medium-L1} & 34.1 & 25.3 & 46.1 & 20.1 & 50.4 & 63.3 & 29.7 & 76.0 & 28.7 & 41.5\textcolor{green!60!black}{\scriptsize{${+0.9}$}}\\
    \rowcolor[HTML]{F2F3F5} \textbf{\MethodName-Medium} & 36.6 & 27.1 & 46.0 & 20.7 & 50.6 & 64.2 & 30.3 & 75.0 & 29.1 & \textbf{42.3\textcolor{green!60!black}{\scriptsize{${+1.7}$}}}\\
    \addlinespace[0.3em]
    GPT-large & 37.2 & 27.4 & 47.0 & 20.2 & 53.6 & 64.6 & 30.3 & 74.9 & 29.8 & 42.8 \\
    GPT-large-PM & 37.2 & 29.4 & 48.4 & 19.5 & 52.4 & 63.8 & 30.6 & 74.3 & 29.8 & 42.8 \\
    ContextLM-Large & 40.7 & 29.9 & 47.9 & 21.1 & 51.3 & 62.9 & 29.8 & 76.6 & 31.0 & 43.9 \\
    \rowcolor[HTML]{F2F3F5} \textbf{\MethodName-Large-L1} & 39.4 & 29.6 & 49.1 & 21.5 & 50.8 & 64.3 & 31.3 & 75.1 & 30.5 & 43.5\textcolor{green!60!black}{\scriptsize{${+0.7}$}}\\
    
    \rowcolor[HTML]{F2F3F5} \textbf{\MethodName-Large} & 41.1 & 30.6 & 48.5 & 19.9 & 52.1 & 65.3 & 31.4 & 77.0 & 30.0 & \textbf{44.0\textcolor{green!60!black}{\scriptsize{${+1.2}$}}}\\
    \addlinespace[0.3em]
    GPT-XL & 39.3 & 29.8 & 49.6 & 20.6 & 51.8 & 65.1 & 31.5 & 77.0 & 30.4 & 43.9 \\
    GPT-XL-PM & 38.7 & 29.5 & 49.2 & 20.0 & 51.2 & 64.8 & 31.5 & 77.0 & 29.9 & 43.5 \\
    ContextLM-XL & 41.9 & 32.3 & 49.9 & 21.4 & 51.9 & 66.1 & 32.3 & 77.7 & 31.2 & 45.0 \\
    \rowcolor[HTML]{F2F3F5} \textbf{\MethodName-xl-L1} & 41.7 & 32.2 & 49.7 & 21.0 & 52.0 & 65.4 & 31.9 & 77.9 & 30.7 & 44.7\textcolor{green!60!black}{\scriptsize{${+0.8}$}}\\
    \rowcolor[HTML]{F2F3F5} \textbf{\MethodName-xl} & 42.9 & 32.2 & 49.9 & 21.9 & 52.1 & 66.8 & 32.4 & 76.8 & 30.6 & \textbf{45.1\textcolor{green!60!black}{\scriptsize{${+1.2}$}}}\\
    
    \midrule
    \midrule
    \textbf{Model} & \multicolumn{2}{c}{OWT} & \multicolumn{2}{c}{Wikitext} & \multicolumn{2}{c}{Lambada OpenAI} & \multicolumn{2}{c}{Lambada Std} & --- & \textbf{Avg PPL} $\downarrow$ \\
    \midrule
    \multicolumn{11}{c}{\textit{Language Modeling (Perplexity)}} \\
    \midrule
    GPT2-Base & \multicolumn{2}{c}{22.04} & \multicolumn{2}{c}{43.99} & \multicolumn{2}{c}{69.29} & \multicolumn{2}{c}{272.43} & --- & 101.94 \\
    GPT2-Base-PM & \multicolumn{2}{c}{21.22} & \multicolumn{2}{c}{42.15} & \multicolumn{2}{c}{60.05} & \multicolumn{2}{c}{273.12} & --- & 99.14 \\
    ContextLM-Base-L1 & \multicolumn{2}{c}{20.68} & \multicolumn{2}{c}{41.45} & \multicolumn{2}{c}{55.22} & \multicolumn{2}{c}{231.41} & --- & 87.19 \\
    \rowcolor[HTML]{F2F3F5} \textbf{\MethodName-Base} & \multicolumn{2}{c}{21.19} & \multicolumn{2}{c}{42.13} & \multicolumn{2}{c}{56.84} & \multicolumn{2}{c}{230.41} & --- & 87.64\textcolor{green!60!black}{\scriptsize{\({+14.30}\)}} \\

    \rowcolor[HTML]{F2F3F5} \textbf{\MethodName-Base} & \multicolumn{2}{c}{\textbf{20.56}} & \multicolumn{2}{c}{\textbf{40.86}} & \multicolumn{2}{c}{\textbf{51.83}} & \multicolumn{2}{c}{\textbf{225.74}} & --- & \textbf{84.75\textcolor{green!60!black}{\scriptsize{\({+17.19}\)}}} \\
    
    \addlinespace[0.3em]

    GPT2-Medium & \multicolumn{2}{c}{17.97} & \multicolumn{2}{c}{34.24} & \multicolumn{2}{c}{35.54} & \multicolumn{2}{c}{122.54} & --- & 52.57 \\
    GPT2-Medium-PM & \multicolumn{2}{c}{17.67} & \multicolumn{2}{c}{33.59} & \multicolumn{2}{c}{32.91} & \multicolumn{2}{c}{126.03} & --- & 52.55 \\
    ContextLM-Medium & \multicolumn{2}{c}{17.03} & \multicolumn{2}{c}{32.08} & \multicolumn{2}{c}{28.06} & \multicolumn{2}{c}{95.06} & --- & 43.06 \\
    \rowcolor[HTML]{F2F3F5} \textbf{\MethodName-Medium-L1} & \multicolumn{2}{c}{17.14} & \multicolumn{2}{c}{32.42} & \multicolumn{2}{c}{31.29} & \multicolumn{2}{c}{100.50} & --- & 43.34\textcolor{green!60!black}{\scriptsize{\({+9.23}\)}}  \\

    \rowcolor[HTML]{F2F3F5} \textbf{\MethodName-Medium} & \multicolumn{2}{c}{\textbf{16.81}} & \multicolumn{2}{c}{\textbf{31.53}} & \multicolumn{2}{c}{\textbf{26.96}} & \multicolumn{2}{c}{\textbf{86.75}} & --- & \textbf{40.51\textcolor{green!60!black}{\scriptsize{\({+12.06}\)}}} 
    \\
    
    \addlinespace[0.3em]
    GPT2-Large & \multicolumn{2}{c}{16.01} & \multicolumn{2}{c}{30.03} & \multicolumn{2}{c}{24.58} & \multicolumn{2}{c}{74.76} & --- & 36.35 \\
    GPT2-Large-PM & \multicolumn{2}{c}{15.89} & \multicolumn{2}{c}{29.71} & \multicolumn{2}{c}{23.50} & \multicolumn{2}{c}{62.45} & --- & 32.89 \\
    ContextLM-Large & \multicolumn{2}{c}{15.41} & \multicolumn{2}{c}{28.82} & \multicolumn{2}{c}{20.89} & \multicolumn{2}{c}{62.09} & --- & 31.80 \\
    \rowcolor[HTML]{F2F3F5} \textbf{\MethodName-Large-L1} & \multicolumn{2}{c}{15.47} & \multicolumn{2}{c}{29.66} & \multicolumn{2}{c}{21.55} & \multicolumn{2}{c}{61.52} & --- & 32.05\textcolor{green!60!black}{\scriptsize{\({+4.30}\)}} \\ 
    \rowcolor[HTML]{F2F3F5} \textbf{\MethodName-Large} & \multicolumn{2}{c}{\textbf{15.20}} & \multicolumn{2}{c}{\textbf{28.52}} & \multicolumn{2}{c}{\textbf{19.68}} & \multicolumn{2}{c}{\textbf{55.05}} & --- & \textbf{29.61\textcolor{green!60!black}{\scriptsize{\({+6.74}\)}}} 
    \\

    \addlinespace[0.3em]
    GPT2-XL & \multicolumn{2}{c}{14.99} & \multicolumn{2}{c}{27.89} & \multicolumn{2}{c}{20.72} & \multicolumn{2}{c}{55.12} & --- & 29.68 \\
    GPT2-XL-PM & \multicolumn{2}{c}{14.94} & \multicolumn{2}{c}{27.95} & \multicolumn{2}{c}{21.00} & \multicolumn{2}{c}{56.15} & --- & 30.01 \\
    ContextLM-XL & \multicolumn{2}{c}{14.60} & \multicolumn{2}{c}{27.05} & \multicolumn{2}{c}{17.46} & \multicolumn{2}{c}{\textbf{44.62}} & --- & \textbf{25.93} \\
    \rowcolor[HTML]{F2F3F5} \textbf{\MethodName-XL-L1} & \multicolumn{2}{c}{14.51} & \multicolumn{2}{c}{26.83} & \multicolumn{2}{c}{18.47} & \multicolumn{2}{c}{49.09} & --- & 27.23\textcolor{green!60!black}{\scriptsize{\({+2.45}\)}} \\
    \rowcolor[HTML]{F2F3F5} \textbf{\MethodName-XL} & \multicolumn{2}{c}{\textbf{14.40}} & \multicolumn{2}{c}{\textbf{26.80}} & \multicolumn{2}{c}{\textbf{17.26}} & \multicolumn{2}{c}{47.06} & --- & 26.38\textcolor{green!60!black}{\scriptsize{\({+3.30}\)}} \\
    
    \bottomrule
    \end{tabular}
    }
    \label{tab:conceptlm_GPT_results_L1}
\end{table}

\subsection{Insert Depth Ablation}
\label{app:insert_depth_abl}

We insert the concept-level layers to L0, L1, L3, L5, L7, L9, L11, the result are shown in Table~\ref{tab:insert_layers}. We find that inserting the concept layers at the three-quarters (3/4) depth of the model yields superior generalization performance in the 160M model size.

\begin{table}[ht]
    \centering
    \small
    \setlength{\tabcolsep}{4.5pt} 
    \caption{Insert depth ablation. Evaluation of Pythia-160M backbone trained on 30B the Pile tokens, L0 denotes there are 0-token-layers before the Concept-level Module.}
    \renewcommand{\arraystretch}{1.2}
    \resizebox{\textwidth}{!}{
    \begin{tabular}{lccccccccc|l}
    \toprule
    \textbf{Model} & \makecell{Lambada \\ OpenAI} & \makecell{Lambada \\ Std.} & \makecell{ARC-\\E} & \makecell{ARC-\\C} & \makecell{Wino\\Gd.} & \makecell{PIQA} & \makecell{Hella-\\Swag} & \makecell{SciQ} & \makecell{RACE} & \textbf{Avg Acc} $\uparrow$ \\
    \midrule
    \multicolumn{11}{c}{\textit{Zero-shot Evaluation (Accuracy \%)}} \\
    \midrule
    
    ConceptLM-L0 & 34.2 & 22.4 & 45.2 & 19.0 & 51.0 & 61.5 & 28.1 & 75.1 & 27.9 & \textbf{40.5} \\
    ConceptLM-L1 & 34.0 & 21.8 & 43.5 & 18.8 & 53.0 & 61.3 & 28.3 & 73.3 & 28.6 & 40.3 \\
    ConceptLM-L3 & 32.2 & 21.6 & 42.0 & 19.7 & 51.1 & 61.4 & 28.1 & 71.9 & 28.8 & 39.6 \\
    ConceptLM-L5 & 33.9 & 22.7 & 43.5 & 19.0 & 51.9 & 61.3 & 28.3 & 73.0 & 29.0 & 40.3 \\
    ConceptLM-L7 & 34.0 & 21.7 & 42.9 & 17.5 & 51.6 & 62.5 & 28.2 & 75.2 & 28.7 & 40.3 \\
    ConceptLM-L9 & 34.9 & 24.2 & 43.1 & 19.7 & 50.3 & 61.9 & 27.8 & 74.2 & 27.9 & 40.4 \\
    ConceptLM-L11& 33.9 & 22.3 & 43.3 & 18.7 & 53.6 & 61.5 & 28.3 & 73.8 & 27.9 & 40.3 \\

    \midrule
    \midrule
    \textbf{Model} & \multicolumn{2}{c}{OWT} & \multicolumn{2}{c}{Wikitext} & \multicolumn{2}{c}{Lambada OpenAI} & \multicolumn{2}{c}{Lambada Std} & --- & \textbf{Avg PPL} $\downarrow$ \\
    \midrule
    \multicolumn{11}{c}{\textit{Language Modeling (Perplexity)}} \\
    \midrule

    ConceptLM-L0 & \multicolumn{2}{c}{12.52} & \multicolumn{2}{c}{33.74} & \multicolumn{2}{c}{37.46} & \multicolumn{2}{c}{166.29} & --- &  63.25 \\
    ConceptLM-L1 & \multicolumn{2}{c}{12.64 } & \multicolumn{2}{c}{33.45} & \multicolumn{2}{c}{36.38} & \multicolumn{2}{c}{190.06} & --- &  68.13 \\
    ConceptLM-L3 & \multicolumn{2}{c}{12.65} & \multicolumn{2}{c}{33.70} & \multicolumn{2}{c}{38.94} & \multicolumn{2}{c}{163.07} & --- &  62.09 \\
    ConceptLM-L5 & \multicolumn{2}{c}{12.66} & \multicolumn{2}{c}{33.62} & \multicolumn{2}{c}{38.09} & \multicolumn{2}{c}{158.20} & --- &  60.64 \\
    ConceptLM-L7 & \multicolumn{2}{c}{12.62} & \multicolumn{2}{c}{33.49} & \multicolumn{2}{c}{35.86} & \multicolumn{2}{c}{175.37} & --- &  64.34 \\
    ConceptLM-L9 & \multicolumn{2}{c}{12.64} & \multicolumn{2}{c}{33.72} & \multicolumn{2}{c}{35.49} & \multicolumn{2}{c}{138.28} & --- &  \textbf{55.04} \\
    ConceptLM-L11 & \multicolumn{2}{c}{18.67} & \multicolumn{2}{c}{32.26} & \multicolumn{2}{c}{36.96} & \multicolumn{2}{c}{150.88} & --- &  59.79 \\
    \bottomrule
    \end{tabular}
    }
    \label{tab:insert_layers}
\end{table}

\subsection{VQ Module Setting}
\label{app:VQ_discussion}

In our early experiments, we employ a vanilla VQ module without the SimVQ trick. We observe that the codebooks frequently collapsed when the concept layers are inserted before the first layer. The training stability is only achieved by applying complex transformations to the codebook through non-linear activation functions. So we keep the 2-MLP layers and ReLU activation setting.

We hypothesize that this collapse occurs because: A simple linear projection (or no projection) lacks the capacity to balance the complex representations with the discrete latent space, causing most inputs to map to a few specific codebook entries. The introduction of non-linear activations redistributes the representations, preventing the model from converging to a few codes and significantly increasing codebook utilization.

Across all our experimental configurations in this paper, the VQ codebook utilization consistently reached nearly 100\%, demonstrating that our design effectively mitigates the common issue of codebook collapse. We provide an early experiment to test the codebook usage even with a large codebook size setting $S=12, N=1024$.

\begin{figure}[htb]
    \centering
    \includegraphics[width=\linewidth]{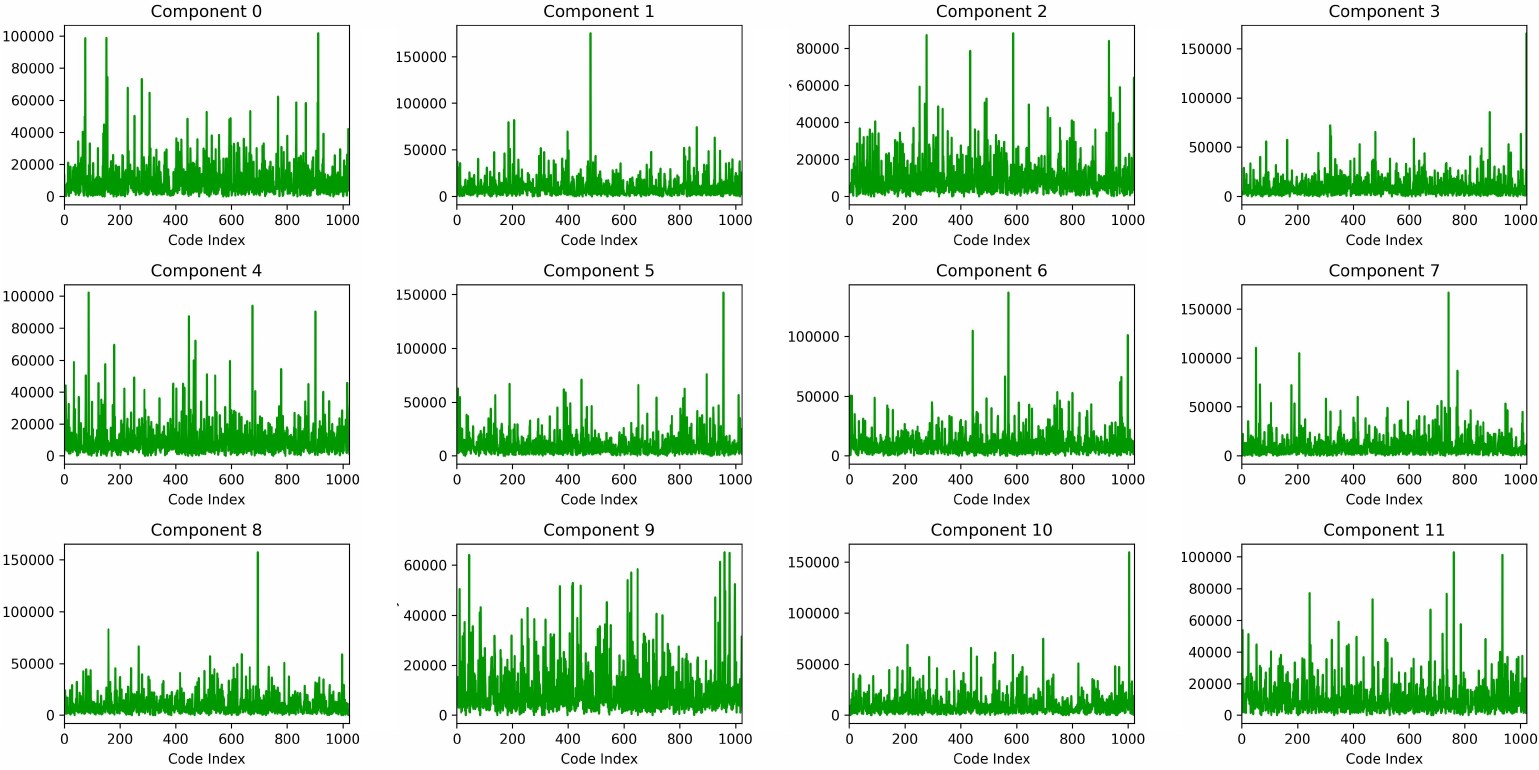}
\caption{Codebook usage. The figure shows the usage frequency of each code entry across 12 codebooks, each containing 1,024 entries, evaluated on the OpenWebText test set.}
\label{fig:codebook_usage}
\end{figure}

\subsection{VQ Ablation}

We provide the experimental results for the codebook size ($N$) ablation in Table~\ref{tab:VQ_N} and the experimental results for the codebook number ($S$) ablation in Table~\ref{tab:VQ_S}. 

The results show that for $N$, adjusting the number of $N$ has little impact on downstream tasks but can significantly affect perplexity (PPL). While a smaller $N$ is easier to predict, a larger $N$ can encode more fine-grained information; achieving optimal results requires a trade-off between the two factors. 

For $S$, appropriately increasing the value reduces the segment size, leading to better fitting and improved downstream performance. However, if $S$ is too large, the fixed-width concept-layer Transformers may struggle to accurately predict the concepts. However, it should be noted that these experiments are conducted with a hidden size of $768$. Further experimentation is required to determine whether larger models can support increased values for $N$ and $S$.

Disregarding the context, for a token vocabulary of size $V$ (vocab\_size) and a chunk size $k$, the representation space of a concept is $V^k$. For a concept vocabulary, the representation range is $N^S$. For our Pythia models experiments, $V\approx50000,k=4,N=64,S=12$, $V^k=6.25*10^{18}$, $N^S\approx72*10^{22}$. $N^S>>V^k$.

\begin{table}[ht]
    \centering
    \small
    \setlength{\tabcolsep}{4.5pt} 
    \caption{VQ codebook size ablation. Evaluation of Pythia-160M backbone with 2 concept layers trained on 30B the Pile tokens. We set $S=\mathrm{number \ of \ head}$, and ablate $N$.}
    \renewcommand{\arraystretch}{1.2}
    \resizebox{\textwidth}{!}{
    \begin{tabular}{lccccccccc|l}
    \toprule
    \textbf{Model} & \makecell{Lambada \\ OpenAI} & \makecell{Lambada \\ Std.} & \makecell{ARC-\\E} & \makecell{ARC-\\C} & \makecell{Wino\\Gd.} & \makecell{PIQA} & \makecell{Hella-\\Swag} & \makecell{SciQ} & \makecell{RACE} & \textbf{Avg Acc} $\uparrow$ \\
    \midrule
    \multicolumn{11}{c}{\textit{Zero-shot Evaluation (Accuracy \%)}} \\
    \midrule
    
    $N=8$ & 33.3 & 21.6 & 43.0 & 19.8 & 51.7 & 60.7 & 28.1 & 74.0 & 27.9 &  40.0 \\
    $N=16$ & 33.9 & 21.6 & 44.7 & 18.7 & 52.0 & 60.5 & 28.1 & 76.2 & 27.1 & 40.3 \\
    $N=32$ & 33.2 & 21.5 & 43.1 & 18.3 & 51.7 & 61.7 & 28.4 & 74.9 & 28.4 & 40.2 \\
    $N=64$ & 34.0 & 21.8 & 43.5 & 18.8 & 53.0 & 61.3 & 28.3 & 73.3 & 28.6 & 40.3\\
    $N=128$ & 34.4 & 22.2 & 44.3 & 18.7 & 53.2 & 62.2 & 28.1 & 74.0 & 26.6 & 40.4 \\
    $N=256$ & 33.4 & 21.9 & 43.3 & 19.5 & 51.4 & 60.6 & 27.9 & 75.6 & 29.2 & 40.3 \\

    \midrule
    \midrule
    \textbf{Model} & \multicolumn{2}{c}{OWT} & \multicolumn{2}{c}{Wikitext} & \multicolumn{2}{c}{Lambada OpenAI} & \multicolumn{2}{c}{Lambada Std} & --- & \textbf{Avg PPL} $\downarrow$ \\
    \midrule
    \multicolumn{11}{c}{\textit{Language Modeling (Perplexity)}} \\
    \midrule

    $N=8$ & \multicolumn{2}{c}{12.70} & \multicolumn{2}{c}{33.87} & \multicolumn{2}{c}{39.89 } & \multicolumn{2}{c}{181.06 } & --- &  66.88 \\
    $N=16$ & \multicolumn{2}{c}{ 12.64 } & \multicolumn{2}{c}{33.71} & \multicolumn{2}{c}{ 35.67 } & \multicolumn{2}{c}{170.08 } & --- &  63.03 \\
    $N=32$ & \multicolumn{2}{c}{ 12.62 } & \multicolumn{2}{c}{33.68} & \multicolumn{2}{c}{ 38.51 } & \multicolumn{2}{c}{ 191.97 } & --- &  69.20 \\
    $N=64$ & \multicolumn{2}{c}{12.64 } & \multicolumn{2}{c}{33.45} & \multicolumn{2}{c}{36.38} & \multicolumn{2}{c}{190.06} & --- &  68.13 \\
    $N=128$ & \multicolumn{2}{c}{12.63} & \multicolumn{2}{c}{33.61} & \multicolumn{2}{c}{ 36.71 } & \multicolumn{2}{c}{ 189.30 } & --- & 68.06 \\
    $N=256$ & \multicolumn{2}{c}{12.68} & \multicolumn{2}{c}{33.45} & \multicolumn{2}{c}{ 39.88 } & \multicolumn{2}{c}{ 195.50 } & --- & 70.38 \\
    
    \bottomrule
    \end{tabular}
    }
    \label{tab:VQ_N}
\end{table}

\begin{table}[ht]
    \centering
    \small
    \setlength{\tabcolsep}{4.5pt} 
    \caption{VQ codebook size ablation. Evaluation of Pythia-160M backbone with 2 concept layers trained on 30B the Pile tokens. We set $N=64$, and ablate $S$, the following $\mathrm{head\ num}*4$ denotes $S = \mathrm{head\ num}*4$ .}
    \renewcommand{\arraystretch}{1.2}
    \resizebox{\textwidth}{!}{
    \begin{tabular}{lccccccccc|l}
    \toprule
    \textbf{Model} & \makecell{Lambada \\ OpenAI} & \makecell{Lambada \\ Std.} & \makecell{ARC-\\E} & \makecell{ARC-\\C} & \makecell{Wino\\Gd.} & \makecell{PIQA} & \makecell{Hella-\\Swag} & \makecell{SciQ} & \makecell{RACE} & \textbf{Avg Acc} $\uparrow$ \\
    \midrule
    \multicolumn{11}{c}{\textit{Zero-shot Evaluation (Accuracy \%)}} \\
    \midrule
    
    $\mathrm{head\ num}*4$ & 34.0 & 21.7 & 43.3 & 19.8 & 51.3 & 62.1 & 28.0 & 74.2 & 27.9 & 40.3 \\
    $\mathrm{head\ num}*2$ & 34.8 & 22.7 & 43.8 & 18.7 & 51.8 & 61.4 & 28.3 & 76.5 & 28.0 & 40.7 \\
    $\mathrm{head\ num}*1$ & 34.0 & 21.8 & 43.5 & 18.8 & 53.0 & 61.3 & 28.3 & 73.3 & 28.6 & 40.3\\
    $\mathrm{head\ num}/2$ & 33.7 & 22.1 & 43.2 & 20.1 & 51.1 & 61.9 & 28.0 & 74.3 & 29.5 & 40.4 \\
    $\mathrm{head\ num}/3$ & 34.2 & 21.5 & 45.2 & 19.5 & 51.8 & 62.0 & 28.2 & 72.2 & 28.3 & 40.3 \\
    $\mathrm{head\ num}/4$ & 33.0 & 21.0 & 45.3 & 18.8 & 50.5 & 61.0 & 28.3 & 74.4 & 26.8 & 39.9 \\
    $\mathrm{head\ num}/6$ & 33.1 & 21.2 & 42.9 & 18.9 & 51.8 & 60.9 & 28.1 & 71.1 & 27.9 & 39.5 \\

    \midrule
    \midrule
    \textbf{Model} & \multicolumn{2}{c}{OWT} & \multicolumn{2}{c}{Wikitext} & \multicolumn{2}{c}{Lambada OpenAI} & \multicolumn{2}{c}{Lambada Std} & --- & \textbf{Avg PPL} $\downarrow$ \\
    \midrule
    \multicolumn{11}{c}{\textit{Language Modeling (Perplexity)}} \\
    \midrule

    $\mathrm{head\ num}*4$ & \multicolumn{2}{c}{12.64} & \multicolumn{2}{c}{33.57} & \multicolumn{2}{c}{38.34 } & \multicolumn{2}{c}{205.29 } & --- &  72.46 \\
    
    $\mathrm{head\ num}*2$ & \multicolumn{2}{c}{ 12.62 } & \multicolumn{2}{c}{33.56} & \multicolumn{2}{c}{36.48 } & \multicolumn{2}{c}{165.45 } & --- &  62.03 \\
    
    $\mathrm{head\ num}*1$ & \multicolumn{2}{c}{12.64 } & \multicolumn{2}{c}{33.45} & \multicolumn{2}{c}{36.38} & \multicolumn{2}{c}{190.06} & --- &  68.13 \\
    
    $\mathrm{head\ num}/2$ & \multicolumn{2}{c}{12.64 } & \multicolumn{2}{c}{33.57} & \multicolumn{2}{c}{37.97} & \multicolumn{2}{c}{164.35} & --- & 62.09  \\
    
    $\mathrm{head\ num}/3$ & \multicolumn{2}{c}{ 12.67 } & \multicolumn{2}{c}{33.73} & \multicolumn{2}{c}{ 37.58 } & \multicolumn{2}{c}{ 184.04 } & --- & 67.00  \\
    
    $\mathrm{head\ num}/4$ & \multicolumn{2}{c}{ 12.68 } & \multicolumn{2}{c}{33.86} & \multicolumn{2}{c}{ 40.28 } & \multicolumn{2}{c}{ 186.71 } & --- & 68.38 \\

    $\mathrm{head\ num}/6$ & \multicolumn{2}{c}{ 12.66 } & \multicolumn{2}{c}{33.82} & \multicolumn{2}{c}{37.44 } & \multicolumn{2}{c}{170.22 } & --- & 63.54 \\

    \bottomrule
    \end{tabular}
    }
    \label{tab:VQ_S}
\end{table}

\subsection{Chunk Size Ablation}

We provide our experimental results for the chunk size ablation. Note that a larger chunk size introduces smaller layer computational cost, showing a clear trade-off between computational costs and model capability. We show the results in Table~\ref{tab:chunk_size}.

\begin{table}[ht]
    \centering
    \small
    \setlength{\tabcolsep}{4.5pt} 
    \caption{Chunk size ablation. Evaluation of Pythia-160M backbone trained on 30B the Pile tokens, Chunk-2 denotes chunk size = 2 ($k=2$).}
    \renewcommand{\arraystretch}{1.2}
    \resizebox{\textwidth}{!}{
    \begin{tabular}{lccccccccc|l}
    \toprule
    \textbf{Model} & \makecell{Lambada \\ OpenAI} & \makecell{Lambada \\ Std.} & \makecell{ARC-\\E} & \makecell{ARC-\\C} & \makecell{Wino\\Gd.} & \makecell{PIQA} & \makecell{Hella-\\Swag} & \makecell{SciQ} & \makecell{RACE} & \textbf{Avg Acc} $\uparrow$ \\
    \midrule
    \multicolumn{11}{c}{\textit{Zero-shot Evaluation (Accuracy \%)}} \\
    \midrule
    
    Chunk-2 & 35.6 & 22.6 & 44.7 & 19.5 & 53.0 & 60.8 & 28.1 & 75.9 & 28.2 & \textbf{40.7} \\

    Chunk-4 & 34.0 & 21.8 & 43.5 & 18.8 & 53.0 & 61.3 & 28.3 & 73.3 & 28.6 & 40.3 \\

    Chunk-8 & 33.7 & 22.5 & 43.5 & 19.2 & 52.3 & 61.4 & 27.3 & 73.0 & 28.6 & 40.2 \\

    Chunk-16 & 33.4 & 21.9 & 42.8 & 19.7 & 50.7 & 60.8 & 27.0 & 74.2 & 28.6 & 39.8 \\

    \midrule
    \midrule
    \textbf{Model} & \multicolumn{2}{c}{OWT} & \multicolumn{2}{c}{Wikitext} & \multicolumn{2}{c}{Lambada OpenAI} & \multicolumn{2}{c}{Lambada Std} & --- & \textbf{Avg PPL} $\downarrow$ \\
    \midrule
    \multicolumn{11}{c}{\textit{Language Modeling (Perplexity)}} \\
    \midrule

    Chunk-2 & \multicolumn{2}{c}{12.59 } & \multicolumn{2}{c}{33.52} & \multicolumn{2}{c}{34.89} & \multicolumn{2}{c}{158.19} & --- &  \textbf{59.80} \\

    Chunk-4 & \multicolumn{2}{c}{12.64 } & \multicolumn{2}{c}{33.45} & \multicolumn{2}{c}{36.38} & \multicolumn{2}{c}{190.06} & --- &  68.13 \\

    Chunk-8 & \multicolumn{2}{c}{12.67 } & \multicolumn{2}{c}{33.82} & \multicolumn{2}{c}{37.35} & \multicolumn{2}{c}{179.41} & --- &  65.81 \\

    Chunk-16 & \multicolumn{2}{c}{12.73 } & \multicolumn{2}{c}{34.09} & \multicolumn{2}{c}{38.73} & \multicolumn{2}{c}{193.26} & --- &  69.70 \\

    \bottomrule
    \end{tabular}
    }
    \label{tab:chunk_size}
\end{table}

\end{document}